\documentclass[letterpaper]{article} 
\usepackage{aaai2025_pkg/aaai25} 
\usepackage{times}  
\usepackage{helvet}  
\usepackage{courier}  
\usepackage[hyphens]{url}  
\usepackage{graphicx} 
\usepackage{natbib}  
\usepackage{caption} 
\usepackage{algorithm}
\usepackage{algorithmic}
\usepackage{amsmath}

\usepackage{amssymb}
\usepackage{amsfonts}
\usepackage{dsfont}
\usepackage{color}
\usepackage{xcolor}
\usepackage{tcolorbox}
\usepackage{tabularx}
\usepackage{multirow}
\usepackage{makecell}
\usepackage{booktabs}
\usepackage{mdframed,lipsum}
\usepackage{newfloat}
\usepackage{listings}
\DeclareCaptionStyle{ruled}{labelfont=normalfont,labelsep=colon,strut=off} 
\floatstyle{ruled}
\newfloat{listing}{tb}{lst}{}
\floatname{listing}{Listing}
\title{RLPF: Reinforcement Learning from Prediction Feedback for User Summarization with LLMs}
\author {
    Jiaxing Wu,
    Lin Ning, Luyang Liu, Harrison Lee, Neo Wu, Chao Wang, \\ Sushant Prakash,
    Shawn O'Banion, Bradley Green, Jun Xie \\[0.001cm]
}
\affiliations {
Google DeepMind \\
\texttt{jxwu@google.com}
}
\usepackage{bibentry}

\begin{document}

\maketitle
\begin{abstract}

LLM-powered personalization agent systems employ Large Language Models (LLMs) to predict users' behavior from their past activities. However, their effectiveness often hinges on the ability to effectively leverage extensive, long user historical data due to its inherent noise and length of such data. Existing pretrained LLMs may generate summaries that are concise but lack the necessary context for downstream tasks, hindering their utility in personalization systems. To address these challenges, we introduce \textbf{R}einforcement \textbf{L}earning from \textbf{P}rediction \textbf{F}eedback (\textbf{RLPF}). RLPF fine-tunes LLMs to generate concise, human-readable user summaries that are optimized for downstream task performance. By maximizing the usefulness of the generated summaries, RLPF effectively distills extensive user history data while preserving essential information for downstream tasks. Our empirical evaluation demonstrates significant improvements in both extrinsic downstream task utility and intrinsic summary quality, surpassing baseline methods by up to 22\% on downstream task performance and achieving an up to 84.59\% win rate on Factuality, Abstractiveness, and Readability. RLPF also achieves a remarkable 74\% reduction in context length while improving performance on 16 out of 19 unseen tasks and/or datasets, showcasing its generalizability. This approach offers a promising solution for enhancing LLM personalization by effectively transforming long, noisy user histories into informative and human-readable representations.

\end{abstract}

%

\section{Introduction}

\begin{figure*}[!t]
\centering
    \includegraphics[width=0.99 \textwidth]{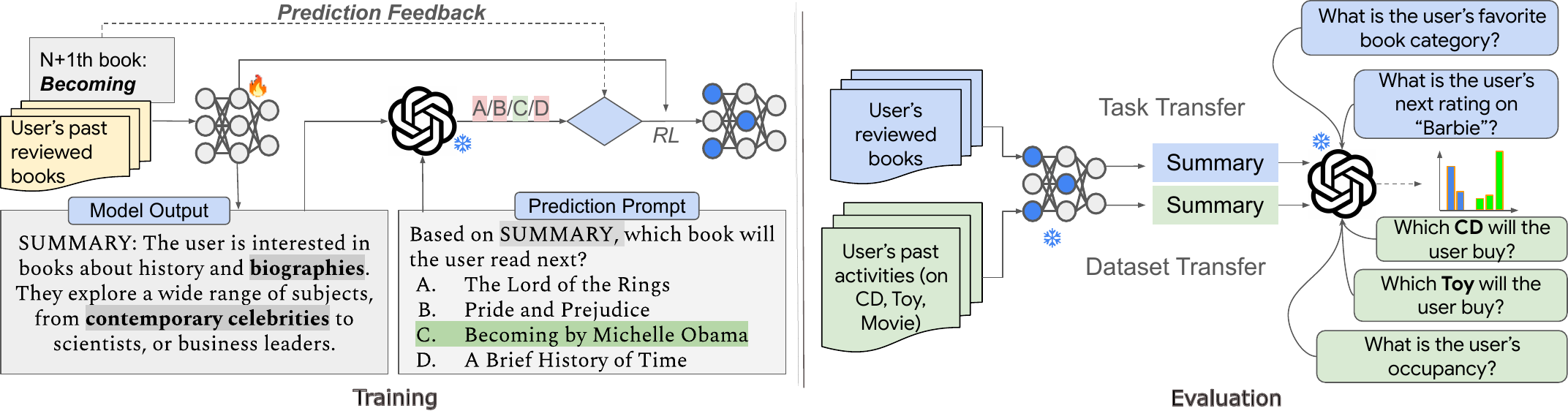}
    \vspace{-0.08in}
    \caption{\textbf{Overview of RLPF}. Left: Training process of RLPF, in which future activity will be used towards reward computation. Right: We assess RLPF on unseen downstream prediction tasks to demonstrate its generalizability and adaptability.}
    \vskip -0.1in
    \label{fig:overview}
\end{figure*}

Large Language Models (LLMs) have shown great promise for personalized prediction by leveraging historical activity data~\cite{liu_chatgptgoodrecommender_2023,lyu_llmrec_2024,li_exploringupperlimitstextbased_2023}. However, the inherent noise and length of user data pose obstacles to their effective utilization in LLM-powered systems. 

Natural language user summaries offer several advantages over using raw user activity data. First, they improve inference efficiency over using raw user data due to their compact nature. Second, they offer the potential to improve performance on downstream tasks by distilling user activities and reducing noise. Representing user context through natural language also offers several advantages over embedding-based representations. User representations in the natural language space are reusable across any LLM for downstream tasks without needing to re-train the LLM. In addition, natural language summaries are interpretable and editable, offering users more scrutability and control over their personalized experiences.

Generating user summaries is inherently challenging because user activities lack a ground-truth summary, and their quality is subjective and difficult to define. Existing techniques share a common shortfall: they offer no guarantee that generated summaries will support downstream personalization tasks—a critical function. Each approach also has unique drawbacks. Heuristic methods that extract subsets of activities fail to capture the breadth of user preferences and often produce less readable results. While prompt engineering is popular, pretrained models are not tailored to user data, and crafting effective prompts is both time-consuming and unscalable. Supervised fine-tuning is impractical due to nonexistent training datasets and the privacy concerns associated with collecting such data. Finally, RLHF or RLAIF methods rely on human or AI evaluators, but their judgments remain subjective without standardized criteria.


To overcome the challenges of generating natural language user summaries, we propose \textbf{RLPF}: \textbf{R}einforcement \textbf{L}earning from \textbf{P}rediction \textbf{F}eedback (illustrated in Figure~\ref{fig:overview}), which includes three components:
\begin{itemize}
    \item \textbf{Summarization Model}: A model is trained to generate succinct user summaries from raw activity data.
    \item \textbf{Prediction-based Reward Model}:  To compute a reward, we measure the effectiveness of the generated summaries in downstream prediction tasks.
    \item \textbf{Feedback Loop}: The reward is then used to update the summarization model with RL, with an additional reward to encourage shorter lengths. This feedback loop guides the summarization model to continuously refine its ability to produce summaries that are not only concise but also highly effective for their intended applications.
\end{itemize}

RLPF offers a win-win solution: it enables the creation of high-quality user summaries without the need for resource-intensive and potentially privacy-compromising human intervention. By directly optimizing the summarization process for downstream prediction performance, we ensure that the generated summaries are both compact and directly relevant to the tasks they are meant to support. Furthermore, compared to prevailing Reinforcement Learning (RL) approaches relying on feedback from a dedicated trained reward LLM~\cite{ouyang_rlhf_2022,bai2022constitutionalai,harrison_rlaif_2024,yang2023rlcd}, RLPF eliminates the overhead of training a separate reward model.

Through extensive experiments on four public datasets grounded in real-world user interactions - MovieLens 2015 and 2003~\cite{Harper2015MovieLens}, Amazon Review~\cite{amazon_data_2016}, and Google Local Review~\cite{yan2022personalized}, we demonstrate that RLPF summaries outperform baselines in terms of predictive power on both seen and unseen tasks, as well as on intrinsic quality evaluations.

Our contributions are four-fold:
\begin{itemize}
    \item We introduce the novel task of generating natural language user summaries for user modeling and personalization systems. This offers an interpretable alternative to traditional embedding-based representations and allows utilization by arbitrary LLMs without further training.
    \item We introduce RLPF, a novel and easy-to-implement method for training user summarizers. RLPF eliminates the need for reference summaries or hand-crafted prompts, while safeguarding user privacy.
    \item We demonstrate that RLPF summaries outperform baselines on both the training task and unseen tasks across four datasets and domains.
    \item We evaluate RLPF summaries intrinsically and find significant improvements in factuality, abstractiveness, and readability.
\end{itemize}
\section{Methodology}

\subsection{Problem Statement}
Consider a set of users $\mathcal{U} = \{u_i\}_{i=1}^\mathcal{M}$, where each user $i$ has an associated chronologically ordered sequence of interactions, denoted as $\{v_i^1, v_i^2, ..., v_i^N\}$. Each $v_i^j$ within this sequence (where $1 \leq j \leq N$) comprises one or more textual features that describe a specific item, such as the titles or ratings of movies watched by the user. For each user $i$, we concatenate all of their interactions $\{v_i^j\}_{j=1}^N$ into a single string to form the user context $u_i$.

A summarizer model $\pi_\theta$ takes as input the user context and generates a summary $s_i = \pi_\theta(u_i)$. The summary is then provided to off-the-shelf LLM to produce a prediction $\hat{y}_i = \mathcal{P}(s_i)$ for a specific downstream task.  We optimize $\pi_\theta$ to generate summaries $\{s_i\}_{i=1}^\mathcal{M}$ that minimize the expected error between the predictions $\{\hat{y}_i\}_{i=1}^\mathcal{M}$ and the ground truth task labels $\{y_i\}_{i=1}^\mathcal{M}$.

\subsection{Reinforcement Learning from Prediction Feedback}
In the context of RL, we formulate summary generation as a Contextual Markov Decision Process (CMDP). In this framework, the state encompasses both the input text and the partially generated summary, while actions correspond to the selection of tokens from the entire vocabulary. At each step, the policy model maps these states to probability distributions over the vocabulary, facilitating autoregressive token selection. This selection process is guided by the current context and the overarching objective of maximizing cumulative rewards. 

Within this RL framework, we formalize RLPF in the context of user summarization as follows:
\begin{itemize}
\item \textbf{State:} The set of user contexts $\mathcal{U} = \{u_i\}_{i=1}^\mathcal{M}$, where each $u_i$ is a single string representing the textual features of a user's $N$ past activities.
\item \textbf{Action:} The set of user summaries $S = \{s_i\}_{i=1}^\mathcal{M}$ generated based on the corresponding user contexts. 
\item \textbf{Policy Model:} The summarizer model, denoted by $\pi_\theta$, which maps user contexts (states) to user summaries (actions): $\pi(u_i; \theta) \rightarrow s_i$.
\item \textbf{Reward:} We leverage a frozen, pre-trained LLM to generate predictions $\mathcal{P}(s_i)$ for one or more specified tasks based on user summaries $s_i$. Then a scalar reward value is computed by comparing the prediction $\mathcal{P}(s_i)$ with its corresponding ground truth label $y_i$ of the specific task.

\end{itemize}

The objective of RLPF is to learn a policy $\pi^*$ that maximizes the expected cumulative reward:
\[\pi^* = \arg \max_{\pi} \mathbb{E}_{u_i \sim \mathcal{U}} [r(\pi(u_i; \theta))]\]

\subsubsection{Reward Computation} RLPF provides the flexibility to leverage any task for reward derivation, tailored to specific downstream application requirements. Moreover, it seamlessly accommodates the combination of rewards from multiple tasks if needed. Our implementation leveraged future activity prediction as the sole task for generating reward signals. This approach demonstrated strong generalization and transferability to unseen tasks, as detailed in the Results section. This underscores the convenience and efficiency of RLPF by eliminating the need for extensive, complex model training and overhead. Further results using alternative reward tasks, along with guidelines for task selection, are provided in the Appendix \ref{app:task_selection}.

For each user $i$, summary reward $r(s_i)$ is as follows:

\[ r(s_i) = r^{pred}(s_i, \: y_i) + w \cdot r^{len}(s_i) \]
where $r^{pred}(.)$ is the prediction feedback reward, $r^{len}(.)$ is the length reward, and $w$ is a weight that controls the balance between the two terms.

\emph{Prediction Feedback Reward:} Recall that each user context $u_i$ consists of the textual features of $N$ past user activities. We employ the subsequent  ($N+1$)-th activity (e.g., watched movie title etc.) as the ground truth label $y_i$ for predicting the future activity. 
Given the user summary $s_i$,  we calculate a binary reward by comparing the LLM’s prediction based on $s_i$ to the actual future activity $v_i^{N+1}$:
\vspace{-0.1in}


\[
r^{pred}(s_i, \:y_i) = \mathds{1}(\mathcal{P}(s_i) = y_i), \: \textit{where  } y_i = v_i^{N+1}
\]


However, since the reward model operates in a zero-shot setting, predicting item names with exact matches without any additional context is challenging due to the vast number of possibilities. This hinders the policy model's ability to receive positive feedback and learn effectively.
To tackle this issue, we adopt a multiple-choice approach, providing four answer choices for each summary based prediction, including the ground truth.
The reward model is then prompted to select the correct option from the given choices. Notably, our method is adaptable to any closed-ended question formats. See Appendix~\ref{app:prompt} for full prompts.

\emph{Length Reward:} Furthermore, to promote concise summary generation, we incorporate a length reward:

\[ r^{len}(s_i) = \min[\mathcal{C}, \: \beta*(\mathcal{L} - l_i)] \]

where $l_i$ represents the token length of summary $s_i$, and the hyperparameters $\mathcal{M}$, $\beta$, and $\mathcal{L}$ denote the upper bound, magnitude, and target length of the summary, respectively. We set the target length to the average length of Zero Shot summaries in our experiments. Please refer to the Appendix~\ref{app:hyperparameter} for specific values of these variables.


\subsubsection{Training Process}
The absence of reference summaries prevents the application of supervised fine-tuning to either the policy or reward model. 
Unlike the standard RLHF pipeline, which sequentially involves supervised fine-tuning, reward modeling, and policy optimization, RLPF directly optimizes the policy in a single RL training step. By leveraging LLMs’ inherent zero-shot summarization and prediction capabilities, RLPF eliminates the need for intricate prompt engineering, generating feedback for the RL process based on predicted future activities.
While RLPF is not tied to any specific RL algorithm, we utilize REINFORCE~\cite{williams1992reinforce} with a baseline to update the policy model given that it is simpler yet still effective for our tasks. Both policy and value models are initialized from a frozen model.

To preserve the LLM's original summarization capability and mitigate reward hacking, we introduce a KL divergence term between the current policy $\pi_\theta$ and the initial policy $\pi_{init}$. Consequently, the policy parameters are updated according to the following rule:
\[\theta \leftarrow \theta + [(1 - \alpha) \nabla_{\theta} \mathbb{E} [r_i] - \alpha \mathbb{E}[\nabla_{\theta} KL(\pi_\theta|| \pi_{init})]]\]
where $\alpha$ is a hyperparameter controlling the balance between the reward maximization and policy regularization.

\section{Experimental Details}
\vspace{-0.1in}
\subsection{Dataset}
We conduct experiments on four public datasets grounded in real-world user interactions, encompassing product reviews, movie watching behavior, and location data. We perform training on \textit{Amazon Books}~\cite{amazon_data_2016},  \textit{Google Local Review}~\cite{yan2022personalized}, \textit{MovieLens 2015}\cite{Harper2015MovieLens}. Additionally, we utilized another four \textit{Amazon Review} datasets with different product categories, as well as \textit{MovieLens 2003}, which features distinct users and movie catalogs compared to \textit{MovieLens 2015}). See appendix~\ref{app:dataset} for dataset details.

\subsubsection{Data Generation}
For each user's interaction data, presented as a chronologically ordered list of activities $u_i \in \mathcal{U}$, we randomly select one item as the target for future activity prediction, denoted as $y_i$. We utilize the $N$ activities preceding this target as the past activities $\{v_i^j\}_{j=1}^N$. 
$v_i^j$ represents an item name and rating pair, where item name correspond to movie title for MovieLens, product name for Amazon Review, and place name $+$ city name for Google Local Review, respectively. 
As previously mentioned, we concatenate $\{v_i^j\}_{j=1}^N$ to construct the user context $u_i$. To prevent label leakage, the last item in each user's data is reserved as the target item in the test set. Unless otherwise specified, we set $N = 50$ in our experiments.

\begin{figure*}[t!]
    \centering
    \includegraphics[width=\textwidth]{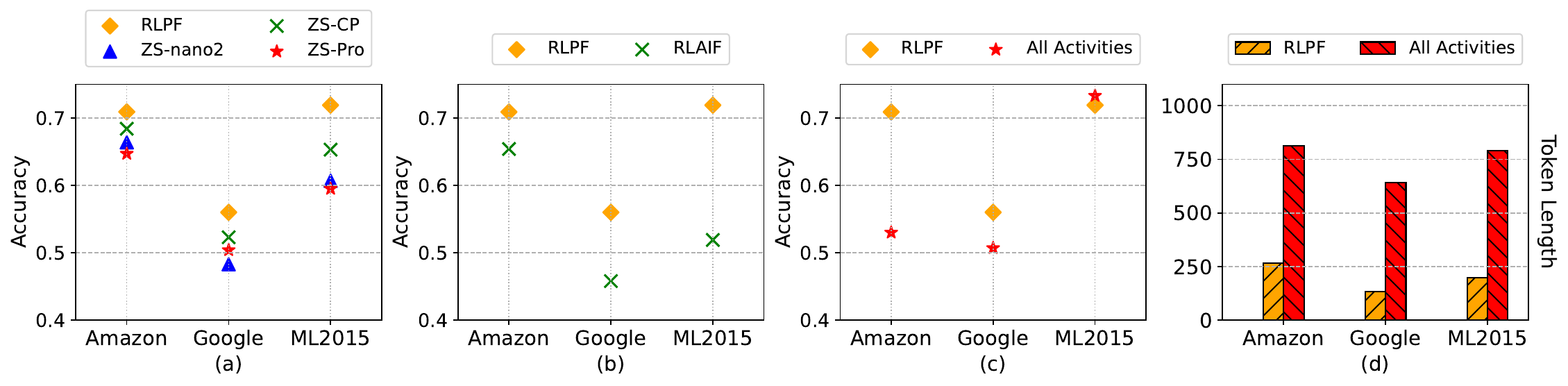}
    \vspace{-0.25in}
    \caption{RLPF summaries consistently demonstrate superior performance in Future Activity Prediction, surpassing both other summarization techniques and the full user context ("All Activities"), while significantly reducing the required context length. ZS-nano2: Gemini Nano-2 Zero-Shot; ZS-CP: Gemini Nano-2 with Crafted Prompts; ZS-Pro: Gemini Pro Zero-Shot.}
    \vskip -0.1in
    \label{fig:main_result}
\end{figure*}

\subsection{Evaluation Metrics}
\subsubsection{Extrinsic Utility} 
We gauge the predictiveness of the summaries based on their prediction performance in various downstream tasks. 
Extending beyond \emph{Future Activity Prediction} which is used as feedback during training, we incorporated additional tasks of various types to gauge the transferability and generalization capabilities of the generated summaries. These included \textbf{19} tasks include user interest reasoning, history activity retrieval, rating prediction, user demographic prediction and open text review generation. Please refer to Appendix~\ref{app:task} for detailed task definitions as well as their abbreviation used in the paper.

A frozen instruction tuned Gemini 1.0 Pro model was employed to generate predictions for all downstream tasks. Each summary $s_i$ was fed into the model, and the resulting predictions were evaluated against ground truth labels.

\subsubsection{Intrinsic Quality }
To further assess the intrinsic quality of the generated summaries, we utilize automated evaluation to compare summaries before and after training. This assessment focuses on aspects not explicitly covered by downstream task performance, including \textit{Factuality}, \textit{Abstractiveness}, \textit{Readability} and \textit{Overall} quality. For each criterion and overall quality, the Auto Rater compares a pair of summaries, with their relative positions randomly assigned to eliminate potential bias in the evaluation. We harnessed the most powerful model in the Gemini family, Gemini 1.5 Pro~\cite{geminiteam2024gemini15unlockingmultimodal}, as the Auto Rater. See Appendix~\ref{app:autoeval} for the full prompt.

In addition to using Auto Rater, the Appendix \ref{app:fres} provides further results and discussions on employing grounded evaluation metrics to assess factuality and readability.
\subsection{Training Details}
The summarizer model, or policy model $\pi_\theta$, is initialized from Gemini 1.0~\cite{geminiteam2024geminifamilyhighlycapable} Nano-2(instruction tuned) and fine-tuned using RLPF. During training, reward computation ($\mathcal{P}(s_i)$) is performed by a frozen, instruction-tuned Gemini 1.0 Pro model, which predicts future activity based on the generated summary $s_i$. Gemini 1.0 Pro was selected for its optimal balance of performance and inference efficiency. In additiona, we also employed PaLM-2 XS~\cite{anil2023palm2technicalreport} to showcase RLPF's applicability across diverse policy models. 

For all three training datasets (Amazon Books, Google Local Review, and MovieLens 2015), we trained the policy model with a batch size of 64, utilizing the Adafactor optimizer \cite{shazeer2018adafactor} with a learning rate of 1e-6. Training was conducted for 15,000 steps on each dataset, and evaluation was performed on the final checkpoint. More hyper-parameter values are listed in Appendix~\ref{app:hyperparameter}.


\subsection{Baselines}
We compare the performance of user summary generated by RLPF against two categories of baselines: summary-based and activity-based. As the evaluator model makes zero-shot predictions for all inputs, any performance differences are attributed to the informativeness of the input, assuming consistent prediction capability.

\begin{itemize}
    \item \textbf{Summary-Based Baselines}: We employ frozen instruction tuned or fine-tuned models to generate summaries and assess their downstream performance.
        \begin{itemize}
            \item \textbf{Gemini 1.0 Nano-2 Zero-Shot:}  Uses summaries generated by Gemini 1.0 Nano-2 in a zero-shot manner. This represents the anchor model before RLPF training.
            \item \textbf{Gemini 1.0 Pro Zero-Shot:}  Uses summaries generated by Gemini 1.0 Pro in a zero-shot manner, a larger and more powerful model than the anchor model.
            \item \textbf{Gemini 1.0 Nano-2 Few-Shot:} Uses summaries generated by Gemini 1.0 Nano-2 in a few-shot manner. We provided two examplars in context, where the example summaries are generated by Gemini 1.5 Pro. See full prompts in Appendix~\ref{app:prompt}.
            \item \textbf{Gemini 1.0 Nano-2 with Crafted Prompt:}  Uses summaries from Gemini 1.0 Nano-2, but with custom-designed prompts optimized for downstream tasks. We show the prompt in Appendix~\ref{app:prompt}.
            \item \textbf{RLAIF:} User summaries trained with Direct RLAIF~\cite{harrison_rlaif_2024}, using Gemini 1.0 Nano-2 as the policy model.  The reward signal is derived from scores provided by an LLM (Gemini 1.0 Pro) based on \emph{Coherence}, \emph{Accuracy}, \emph{Coverage}, and \emph{Overall} quality.  Further details on the prompting technique are available in the Appendix~\ref{app:prompt}.
        \end{itemize}
    \item \textbf{Activity-Based Baselines}: The user context $u_i$is directly fed as input to a frozen instruction tuned model(Gemini 1.0 Pro) to generate predictions~\cite{lyu_llmrec_2024}:
        \begin{itemize}
            \item \textbf{First $X$ Activities:} Uses only the earliest X activities ($X < N$) for downstream task predictions, ensuring comparable token length to RLPF summaries.
            \item \textbf{Random $X$ Activities:}  Similar to the above, but selects $X$ activities randomly.
            \item \textbf{Last $X$ Activities:}  Uses the most recent $X$ activities.
            \item \textbf{All Activities:}  Uses the full user context $N$ activities.
        \end{itemize}
\end{itemize}

\section{Results}
\begin{table*}[]
\centering
\small
\begin{tabular}{clllcrccc}
\toprule
\multicolumn{1}{l}{} & \textbf{\begin{tabular}[c]{@{}c@{}}Training\\ Dataset\end{tabular}} & \textbf{\begin{tabular}[c]{@{}c@{}}Evaluation\\ Dataset\end{tabular}} & \textbf{\begin{tabular}[c]{@{}c@{}}Evaluation\\ Task\end{tabular}} & \textbf{0-Shot} & \multicolumn{1}{c}{\textbf{RLAIF}} & \textbf{RLPF} & \textbf{\begin{tabular}[c]{@{}c@{}}vs \\ 0-Shot\end{tabular}} & \multicolumn{1}{c}{\textbf{\begin{tabular}[c]{@{}c@{}}vs \\ RLAIF\end{tabular}}} \\
\midrule
 & MovieLens  2015 & MovieLens  2015 & Fav Genre & 0.774 & 0.776 & \textbf{0.818} & {5.68\%} & 5.48\% \\
 & MovieLens  2015 & MovieLens  2015 & Rating & 0.225 & 0.229 & \textbf{0.232} & {3.11\%} & 1.31\% \\
 & Amazon Books & Amazon Books & Fav Category & 0.594 & \textbf{0.613} & 0.605 & {1.85\%} & -1.27\% \\
 & Amazon Books & Amazon Books & Rating & 0.244 & 0.147 & \textbf{0.255} & {4.51\%} & 73.47\% \\
 & Amazon Books & Amazon Books & Review Gen & 13.52 & \textbf{13.68} & 13.46 & {-0.41\%} & -1.58\% \\
 & Google Local & Google Local & Fav Category & 0.487 & 0.513 & \textbf{0.559} & {14.78\%} & 8.90\% \\
 & Google Local & Google Local & Rating & \textbf{0.118} & \textbf{0.118} & 0.111 & {-5.93\%} & -5.93\% \\
\multirow{-8}{*}{\begin{tabular}[c]{@{}c@{}}Task\\ Transfer\end{tabular}} & Google Local & Google Local & Common City & 0.765 & 0.791 & \textbf{0.901} & {17.73\%} & 13.93\% \\
\midrule
 & MovieLens  2015 & MovieLens  2003 & Future Act & 0.468 & 0.447 & {\textbf{0.509}} & {8.82\%} & 13.93\% \\
 & MovieLens  2015 & Amazon Movies & Future Act & 0.572 & 0.579 & \textbf{0.606} & {5.94\%} & 4.66\% \\
 & Amazon Books & Amazon Movies & Future Act & 0.645 & 0.573 & \textbf{0.663} & {2.73\%} & 15.68\% \\
 & Amazon Books & Amazon CDs & Future Act & 0.397 & 0.447 & \textbf{0.573} & {44.33\%} & 28.22\% \\
 & Amazon Books & Amazon Toys & Future Act & 0.620 & 0.585 & \textbf{0.644} & {3.94\%} & 10.14\% \\
\multirow{-6}{*}{\begin{tabular}[c]{@{}c@{}}Dataset\\ Transfer\end{tabular}} & Amazon Books & Amazon Games & Future Act & 0.688 & 0.631 & \textbf{0.713} & {3.60\%} & 12.90\% \\
\midrule
 & MovieLens  2015 & MovieLens  2003 & Fav Genre & 0.808 & 0.801 & {\textbf{0.843}} & {4.35\%} & 5.26\% \\
 & MovieLens  2015 & MovieLens  2003 & User Age & 0.274 & \textbf{0.341} & 0.246 & {-10.22\%} & -27.86\% \\
 & MovieLens  2015 & MovieLens  2003 & User Gender & 0.723 & \textbf{0.738} & {0.729} & {0.90\%} & -1.15\% \\
 & MovieLens  2015 & MovieLens  2003 & User Occupancy & 0.146 & 0.130 & {\textbf{0.162}} & {11.20\%} & 24.89\% \\
\multirow{-5}{*}{\begin{tabular}[c]{@{}c@{}}Task \&\\ Dataset\\ Transfer\end{tabular}} & MovieLens  2015 & MovieLens  2003 & Rating & 0.228 & 0.224 & {\textbf{0.245}} & {7.50\%} & 9.38\% \\
\bottomrule
\end{tabular}
\vspace{-0.05in}
\caption{RLPF, trained exclusively on future activity prediction, exhibits remarkable transferability and generalization across diverse unseen tasks and datasets. Evaluation metrics: recall@3 for Favorite Genre/Category, Common City, and User Occupancy; ROUGE-Lsum for Review Gen; and accuracy for the remaining tasks.}
\label{table:transfer}
\end{table*}

\subsection{Target Task Performance}
Figure \ref{fig:main_result} compares RLPF performance on the Future Activity Prediction task. Across all three datasets, RLPF demonstrates superior or comparable performance to various summarizers, including crafted prompting, a larger summarizer model, and RLAIF. Overall, RLPF outperforms Nano-2 zero-shot summaries by +13.4\% improvement, and outperforms RLAIF by +22\% on average. Compared to utilizing the full user context (all activities), RLPF achieves an average context length compression of -73.8\% while still exhibiting a +12.4\% performance gain. Further comparisons with other baselines are provided in the Appendix~\ref{app:results}, underscoring exceptional capability of RLPF summaries to capture both short-term and long-term user context information.

For comparison, we conducted supervised fine-tuning of a Gemini 1.0 Pro model on the same task, reaching 94\% accuracy. However, this fine-tuned model exhibited zero performance on other tasks, highlighting its overfitting to the specific training task. Conversely, RLPF showcased remarkable transferability and generalization capabilities, as demonstrated in the subsequent section.

\subsection{Transferability and Generalization}
To evaluate the generalizability and adaptability of RLPF for various personalization agent systems, we conducted a comprehensive transferability assessment across a diverse set of unseen tasks and datasets. As shown in Table~\ref{table:transfer}, RLPF summaries consistently exhibited superior transferability compared to zero-shot and RLAIF baselines, demonstrating improvements in 16 and 14 out of 19 total evaluation cases, respectively. These results highlight RLPF's exceptional transferability and its potential to be effectively applied to a wide range of personalization scenarios, particularly when training data is scarce.

\subsubsection{Task Transfer}
RLPF summaries demonstrated a slight improvement on an unseen retrieval task, common city retrieval on Google Local Review, and performed on par with zero-shot summary on an unseen personalized text generation task, review generation on Amazon Books.
\subsubsection{Dataset and Domain Transfer}
We also evaluated whether an RLPF trained model can generalize to an unseen dataset, either in same domain or a different domain. We used the policy model trained with MovieLens 2015 to generate summaries on MovieLens 2003 and Amazon Movies\&TVs dataset and evaluated future movie prediction with the generated summaries. 
From the results, RLPF model trained on MovieLens 2015, showed improvements on both unseen datasets. Furthermore, the model trained on Amazon Books achieved significant performance gains on Amazon CDs\&Vinyl data, highlighting its strong domain adaptation abilities.
\subsubsection{Task and Dataset Transfer}
Furthermore, we evaluated RLPF model performance on unseen tasks from unseen datasets. RLPF model trained with MovieLens 2015 with future activity prediction showed improvement on MovieLens 2003 dataset in favorite genre prediction and user demographic reasoning. 
\begin{table}[]
\scalebox{0.9} {

\begin{tabular}{cccc}
\toprule
 &  & \multicolumn{2}{c}{\textbf{RLPF Win Rate}} \\
\multirow{-2}{*}{\textbf{Dataset}} & \multirow{-2}{*}{\textbf{Criteria}} & vs Zero-Shot & vs RLAIF \\
\midrule
 & Factuality & 61.32\% & 62.53\% \\
 & Abstractiveness & 62.54\% & 56.09\% \\
 & Readability & 62.42\% & 56.36\% \\
\multirow{-4}{*}{\begin{tabular}[c]{@{}c@{}}MovieLens \\ 2015\end{tabular}} & Overall & 62.47\% & 56.10\% \\
\midrule
 & Factuality & 72.93\% & 40.09\% \\
 & Abstractiveness & 70.14\% & 39.20\% \\
 & Readability & 71.28\% & 35.47\% \\
\multirow{-4}{*}{\begin{tabular}[c]{@{}c@{}}Amazon \\ Books\end{tabular}} & Overall & 70.08\% & 39.17\% \\
\midrule
 & Factuality & 77.58\% & 49.97\% \\
 & Abstractiveness & 84.59\% & 54.56\% \\
 & Readability & 83.73\% & 46.02\% \\
\multirow{-4}{*}{\begin{tabular}[c]{@{}c@{}}Google \\ Local \\ Review\end{tabular}} & Overall & 84.46\% & 54.22\% \\
\bottomrule
\end{tabular}
}
\vspace{-0.05in}
\caption{Intrinsic Evaluation with Auto Rater.}
\vskip -0.05in
\label{table:autoeval}
\end{table}

\subsection{Intrinsic Evaluation}


Table~\ref{table:autoeval} demonstrates that RLPF summaries consistently outperform zero-shot summaries on all three datasets, as evaluated by the automated rater across all criteria: \emph{Factuality}, \emph{Abstractiveness}, and \emph{Readability}, as well as in the \emph{Overall} evaluation.

This finding is noteworthy given that RLPF was trained solely on reward signals from future activity prediction. Despite this focused training, RLPF summaries not only avoid degradation or overfitting to a single goal but also exhibit significant improvements in other crucial aspects. This suggests that when employing RLPF for user summarization, designing explicit reward signals for each criterion, which can be challenging to obtain, may not be necessary. Instead, future activity prediction performance appears to provide correlated and implicit signals for these criteria. Intuitively, to make accurate future activity predictions, a summary needs to be factually consistent and distill key user information. While readability might not be a strict prerequisite for future activity prediction, it's noteworthy that this criterion also correlates with this downstream task.

Interestingly, RLPF's performance on par with RLAIF in this evaluation, even though RLAIF was specifically trained with reward signals more aligned with the intrinsic evaluation criteria, highlights the effectiveness of RLPF.


\subsection{Analysis}
\subsubsection{Alternative Policy Model}
Additionally, we applied RLPF to a policy model initialized from the PaLM-2 XS model, with results presented in Table~\ref{table:palm2}. Mirroring the observations with Gemini 1.0 Nano-2, RLPF summaries based on PaLM-2 XS also exhibited improvements in both the training task (future activity prediction) and the unseen task (favorite genre/category prediction) across all three datasets. A slight drop in performance was noted for favorite genre prediction on the MovieLens 2015 dataset.

\subsubsection{Robustness to Model that Uses Summaries}
To further ensure that RLPF summaries are not overly tailored to the specific reward model used during training, we employed an additional evaluator model PaLM-2 S to assess their performance. As in previous experiments, RLPF summaries were trained using reward signals derived from Gemini 1.0 Pro.  Table~\ref{table:alternative_evaluator} demonstrates that the improvements achieved with RLPF summaries transfer effectively to these different evaluator models, highlighting the generalizability of RLPF summaries across various LLM-powered systems.


\begin{table}[]
\scalebox{0.95}{
\begin{tabular}{cccc}
\toprule
 &  & \multicolumn{2}{c}{\textbf{PaLM-2 XS}} \\
\multirow{-2}{*}{\textbf{Dataset}} & \multirow{-2}{*}{\textbf{Task}} & \textbf{zero-shot} & \textbf{RLPF} \\
\midrule
 & Future Act & 0.638 & \textbf{0.741} \\
\multirow{-2}{*}{\begin{tabular}[c]{@{}c@{}}MovieLens\\ 2015\end{tabular}} & Fav Category & 0.860 & \textbf{0.849} \\
\midrule
 & Future Act & 0.626 & \textbf{0.675} \\
\multirow{-2}{*}{\begin{tabular}[c]{@{}c@{}}Amazon \\ Books\end{tabular}} & Fav Category & 0.557 & \textbf{0.565} \\
\midrule
 & Future Act & 0.502 & \textbf{0.532} \\
\multirow{-2}{*}{\begin{tabular}[c]{c@{}}Google Local \\ Review\end{tabular}} & Fav Category & 0.454 & \textbf{0.477} \\
\bottomrule
\end{tabular}
}
\vspace{-0.05in}
\caption{RLPF with PaLM-2 XS as the policy model.}
\label{table:palm2}
\end{table}

\begin{table}[]
\begin{tabular}{cccc}
\toprule
\textbf{Dataset} & \textbf{Task} & \textbf{zero-shot} & \textbf{RLPF} \\
\midrule
\multirow{2}{*}{\begin{tabular}[c]{@{}c@{}}MovieLens\\ 2015\end{tabular}} & Future Act & 0.578 & \textbf{0.674} \\
 & Fav Category & 0.822 & \textbf{0.840} \\
\midrule
\multirow{2}{*}{\begin{tabular}[c]{@{}c@{}}Amazon \\ Books\end{tabular}} & Future Act & 0.689 & \textbf{0.734} \\
 & Fav Category & 0.543 & \textbf{0.567} \\
\bottomrule
\end{tabular}
\vspace{-0.05in}
\caption{Evaluated using PaLM-2 S, with reward signals derived from Gemini 1.0 Pro during training.}
\label{table:alternative_evaluator}
\vspace{-0.05in}
\end{table}

\subsubsection{Impact of Summary Length}

\begin{figure}[t!]
    \includegraphics[width=0.9\columnwidth]{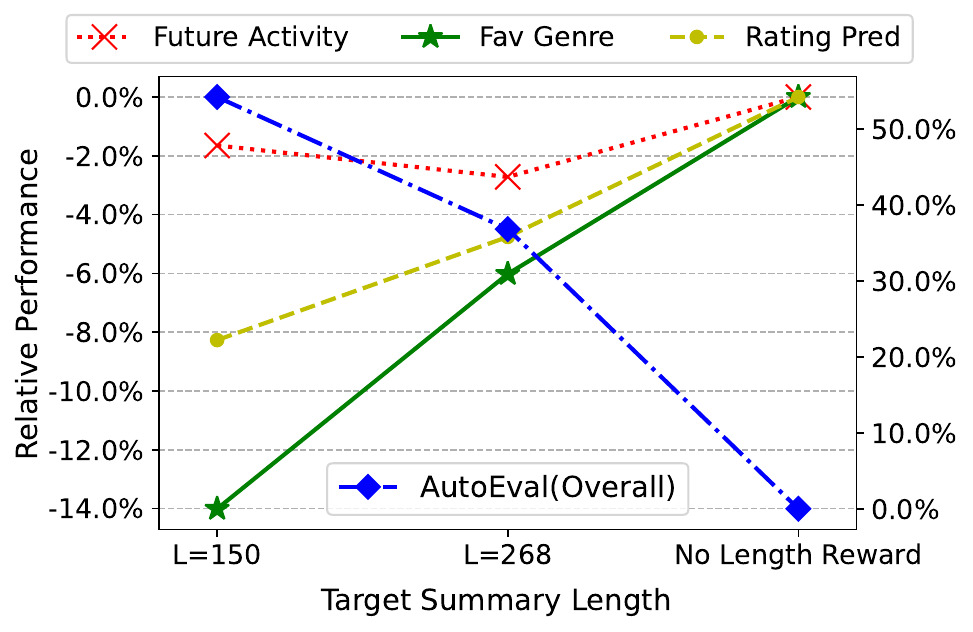}
    \vspace{-0.1in}
    \caption{Impact of Different Target Lengths on MovieLens 2015. Percentage changes are calculated relative to ``No Length Reward'' condition (no maximum length constraint). Data on the right axis pertains to AutoEval, while the left axis corresponds to the remaining tasks.}
    \label{fig:length}
\vspace{-0.05in}
\end{figure}

Figure \ref{fig:length} illustrates our experiments on MovieLens 2015, where we varied the target length($\mathcal{L}$) in the length reward term. Generally, longer summaries led to improved task performance but decreased scores in automated evaluation metrics, suggesting a trade-off between extrinsic utility and intrinsic qualities.

\subsubsection {Robustness to Prompts}
\begin{figure}[t!]
    \includegraphics[width=0.9\columnwidth]{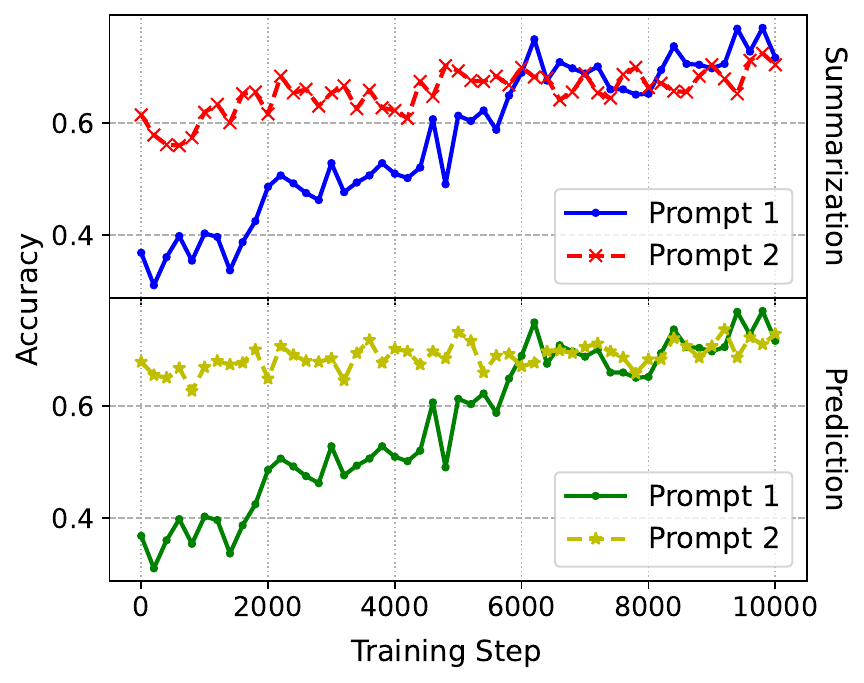}
    \vspace{-0.01in}
    \caption{\textbf{RLPF is robust with various prompts.} Top: Evaluation metric with different prompts for \emph{Summarization}, Bottom: Evaluation metric with different prompts for \emph{Prediction} during reward computation. Prediction task: Future activity prediction on MovieLens 2015.}
    \label{fig:robust}
\vspace{-0.01in}
\end{figure}
We investigated the impact of varying prompts for summary generation and prediction during reward computation. As illustrated in Figure \ref{fig:robust}, task returns converge to a similar level despite initial differences in zero-shot performance, demonstrating the robustness of RLPF to diverse prompts. See full prompts in Appendix~\ref{app:prompt}.

\subsubsection{Qualitative Observation}
In general, zero-shot summaries tend to mimic the structure of the input, which may either be directly copied from the input activities or represent hallucinations (e.g., mentioning popular movies like ``The Godfather'' despite their absence in the user history). After RLPF training, summaries become more coherent and distill user information effectively, though some repetition or hallucination may still occur occasionally. This also explains the better Factuality and Abstractiveness scores from the Automated Evaluation. Although, we noticed both RLPF and RLAIF summaries sometimes exhibit repetitive patterns (e.g., "I am always up for a good recommendation"), while the core content remains user-specific. See the Appendix~\ref{app:example_summary} for example summaries.
\section{Discussion}

\subsubsection{Responsible Deployment}
While RLPF shows promise for enhancing personalization, its use of user data raises privacy and data leakage concerns. Offline training of user summaries and employing a frozen LLM for online serving can mitigate some risks. However, a thorough analysis of potential vulnerabilities is crucial before real-world deployment.

\section{Related Work}
\subsection{Text Summarization}
Leveraging language models to summarize long documents has become prominent. A key distinction between text summarization and user summarization lies in the objectives. The former aims to condense long texts into shorter versions while preserving key information. In contrast, our work seeks to distill implicit user insights and preferences beyond simple extraction from user history.

User summarization poses distinct challenges in model training and evaluation due to the absence of ground truth user summaries. In text summarization, widely-used datasets with reference summaries \cite{hermann2015teaching, narayan2018dont, napoles2012annotated} enable supervised fine-tuning~\cite{cohan-etal-2018,he-ctrlsum-2022,he-z-2023,kryscinski-booksum-2021,roit2023factually} or RL with reward signals comparing generated and reference summaries~\cite{gunasekara2021using}, as well as evaluation metrics with lexical matching~\cite{lin-rouge-2004} or embedding similarity~\cite{zhang2020bertscore}. These methods and metrics are inapplicable to user summarization due to the lack of datasets.
Human evaluation has been used in text summarization~\cite{goyal_newssummarizationevaluation_2023}, but privacy concerns make it impractical for user summarization.

Previous summarization work without reference summaries aligns more with ours. These methods often leverage question-answering (QA)~\cite{durmus-etal-2020,fabbri-qafacteval-2022,deutsch-etal-2021,fabbri_2021} or pre-trained models~\cite{kryscinski-evaluating-2020,goyal-durrett-2020}, relying on the capabilities of QA generation or entailment models. However, no datasets exist for training these models on user activity data. Our work also employs QA and pre-trained LLMs for reward computation, but takes a practical approach by grounding reward signals in real-world personalization questions with answers derived directly from user data, avoiding the need to train additional QA models.

\subsection{User Modeling}

User modeling has benefited significantly from LLM advancements. While existing methods often represent user activity with embeddings~\cite{lin_user_llm_2024,doddapaneni-etal-2024}, our work generates natural language-based user summaries, a more human-readable and reusable alternative.

Previous work on natural language-based user modeling has primarily relied on prompting or fine-tuning for specific downstream tasks~\cite{tallrec_2023,wu_surveylargelanguagemodels_2024,liu_chatgptgoodrecommender_2023,lyu_llmrec_2024,li_exploringupperlimitstextbased_2023,lamp_2024,wang2024usersumbenchbench} or pre-defined user attributes~\cite{rao2023can, ji2023chatgpt,wu2024understandingroleuserprofile}. In contrast, our approach introduces a novel end-to-end training framework for generating user summaries. This method focuses on comprehensive user profiling to support a wide range of downstream tasks, rather than focusing on a single user attribute or characteristic.


\subsection{Reinforcement Learning from AI Feedback}
RL from Human Feedback (RLHF)~\cite{ouyang_rlhf_2022} aligns language models with human values but relies heavily on high-quality human labels. To mitigate this dependency, RL from AI Feedback (RLAIF)\cite{bai2022constitutionalai, yang2023rlcd} utilizes off-the-shelf LLMs to replace human annotations, achieving superior performance on tasks like summarization~\cite{harrison_rlaif_2024}. RLAIF scores summaries directly using an LLM, which introduces subjectivity due to the lack of standardized criteria. In contrast, our approach RLPF uses downstream task performance as the reward signal, enabling direct optimization for improved personalization.



\section{Conclusions}
We introduced RLPF, a novel method to generate human-readable user summaries from raw activity data. RLPF leverages readily available LLMs and downstream task performance as reward signals, overcoming challenges in traditional summarization approaches. Our experiments demonstrate superior performance, context compression, and generalizability across unseen tasks. Future work will extend RLPF to more complex scenarios, additional feedback mechanisms, and broader applications.



\newpage

\bibliography{references.bib}

\onecolumn
\appendix
\newpage

\section{Automated Evaluation Details}
\label{app:autoeval}

Table~\ref{autoeval_prompt} shows the prompt we used for automated evaluation, which assesses the summary qualify from three axes: (1) Factuality, (2) Abstractiveness, and (3) Readability. We also asked auto raters for an overall assessment. The order of the two summaries in the prompt is selected randomly to avoid evaluation bias caused by position. Detailed results are shown in Table \ref{tab:autoeval_zeroshot} and Table \ref{tab:autoeval_rlaif}.

\begin{tcolorbox}
{\fontfamily{cmss}\selectfont
You are an evaluator to judge which user summary is better. The user summary will be used in agent systems for personalization.

You will be provided with two user summaries to compare with, along with the input user activity history that was used to generate those two summaries.

Your job is to evaluate the summaries based on following three criteria:\\

\textbf{1. Factuality: Does the user summary only contain facts supported by the activity history?\\
2. Abstractiveness: Does the user summary capture the key insights of the user based on their activity history in a concise manner, instead of repeating the activity history?\\
3. Readability: Is the user summary fluent and readable?
And based on these, also give an overall evaluation.\\}
~\\
Here is the user activity history (between $<$history$>$ and $<$/history$>$):\\
$<$history$>$
\{input\_activities\}
$<$/history$>$
\\

Here is the first user summary (between $<$summary1$>$ and $<$/summary1$>$):\\
$<$summary1$>$
SUMMARY\_1
$<$/summary1$>$
\\

Here is the second user summary (between $<$summary2$>$ and $<$/summary2$>$):\\
$<$summary2$>$
SUMMARY\_2
$<$/summary2$>$ \\
~\\
For each criteria and the overall comparison:
Respond 0 if the first summary is better;
Respond 1 if two summaries are equally good;
Respond 2 if the second summary is better.

Please structure your response in the following format with no extra explanation.\\
Factuality: Your choice of 0/1/2\\
Abstractiveness: Your choice of 0/1/2\\
Readability: Your choice of 0/1/2\\
Overall: Your choice of 0/1/2 \\
~\\
Response:
}
\end{tcolorbox}
\noindent\begin{minipage}{1.0\textwidth}
\captionof{table}{Evaluation Criteria and Prompt Used for Automated Evaluation.}\label{autoeval_prompt}
\end{minipage}


\begin{table*}[h]
\centering
\begin{tabular}{ccccccc}
\toprule
\textbf{Dataset} & { \textbf{Criteria}} & \multicolumn{1}{c}{{ \textbf{\begin{tabular}[c]{@{}c@{}}0-Shot Win\end{tabular}}}} & \multicolumn{1}{c}{{ \textbf{Equal}}} & \multicolumn{1}{c}{{ \textbf{\begin{tabular}[c]{@{}c@{}}RLPF Win\end{tabular}}}} & \multicolumn{1}{c}{{ \textbf{\begin{tabular}[c]{@{}c@{}}Equal Ratio\end{tabular}}}} & \multicolumn{1}{c}{{ \textbf{\begin{tabular}[c]{@{}c@{}}RLPF Win Rate\end{tabular}}}} \\
\midrule
 & { \textbf{Factuality}} & { 2908} & { 1} & { 4611} & { 0.0\%} & \textbf{ 61.3\%} \\
 & { \textbf{Abstractiveness}} & { 2814} & { 8} & { 4698} & { 0.1\%} & \textbf{ 62.5\%} \\
 & { \textbf{Readability}} & { 2441} & { 1024} & { 4055} & { 13.6\%} & \textbf{ 62.4\%} \\
\multirow{-4}{*}{\begin{tabular}[c]{@{}c@{}}MovieLens\\ 2015\end{tabular}} & { \textbf{Overall}} & { 2822} & { 0} & { 4698} & { 0.0\%} & \textbf{ 62.5\%} \\
\midrule
 & { \textbf{Factuality}} & { 2043} & { 644} & { 5505} & { 7.9\%} & \textbf{ 72.9\%} \\
 & { \textbf{Abstractiveness}} & { 2414} & { 107} & { 5671} & { 1.3\%} & \textbf{ 70.1\%} \\
 & { \textbf{Readability}} & { 2133} & { 764} & { 5295} & { 9.3\%} & \textbf{ 71.3\%} \\
\multirow{-4}{*}{\begin{tabular}[c]{@{}c@{}}Amazon\\ Books\end{tabular}} & { \textbf{Overall}} & { 2426} & { 83} & { 5683} & { 1.0\%} & \textbf{ 70.1\%} \\
\midrule
 & { \textbf{Factuality}} & { 1835} & { 6} & { 6351} & { 0.1\%} & \textbf{ 77.6\%} \\
 & { \textbf{Abstractiveness}} & { 1262} & { 5} & { 6925} & { 0.1\%} & \textbf{ 84.6\%} \\
 & { \textbf{Readability}} & { 1264} & { 425} & { 6503} & { 5.2\%} & \textbf{ 83.7\%} \\
\multirow{-4}{*}{\begin{tabular}[c]{@{}c@{}}Google\\ Local\\ Review\end{tabular}} & { \textbf{Overall}} & { 1273} & { 0} & { 6919} & { 0.0\%} & \textbf{ 84.5\%} \\
\bottomrule
\end{tabular}
\caption{Automated Evaluation Results on Comparing Gemini 1.0 Nano-2 zero shot v.s. RLPF. Note that the total number of examples may fluctuate as we exclude instances where either response doesn't adhere to the guidelines.}
\label{tab:autoeval_zeroshot}
\end{table*}

\begin{table*}[t]
\centering
\begin{tabular}{ccccccc}
\toprule
\textbf{Dataset} & { \textbf{Criteria}} & \multicolumn{1}{c}{{
\textbf{\begin{tabular}[c]{@{}c@{}}RLAIF Win\end{tabular}}}} & \multicolumn{1}{c}{{ \textbf{Equal}}} & \multicolumn{1}{c}{{ \textbf{\begin{tabular}[c]{@{}c@{}}RLPF Win\end{tabular}}}} & \multicolumn{1}{c}{{ \textbf{\begin{tabular}[c]{@{}c@{}}Equal Ratio\end{tabular}}}} & \multicolumn{1}{c}{{ \textbf{\begin{tabular}[c]{@{}c@{}}RLPF Win Rate\end{tabular}}}} \\
\midrule
 & { \textbf{Factuality}} & { 2814} & { 0} & { 4696} & { 0.0\%} & \textbf{ 62.5\%} \\
 & { \textbf{Abstractiveness}} & { 3298} & { 0} & { 4212} & { 0.0\%} & \textbf{ 56.1\%} \\
 & { \textbf{Readability}} & { 3217} & { 138} & { 4155} & { 1.8\%} & \textbf{ 56.4\%} \\
\multirow{-4}{*}{\begin{tabular}[c]{@{}c@{}}MovieLens\\ 2015\end{tabular}} & { \textbf{Overall}} & { 3297} & { 0} & { 4213} & { 0.0\%} & \textbf{ 56.1\%} \\
\midrule
 & { \textbf{Factuality}} & { 4890} & { 23} & { 3272} & { 0.3\%} & { 40.1\%} \\
 & { \textbf{Abstractiveness}} & { 4960} & { 27} & { 3198} & { 0.3\%} & { 39.2\%} \\
 & { \textbf{Readability}} & { 3848} & { 2222} & { 2115} & { 27.1\%} & { 35.5\%} \\
\multirow{-4}{*}{\begin{tabular}[c]{@{}c@{}}Amazon\\ Books\end{tabular}} & { \textbf{Overall}} & { 4972} & { 12} & { 3201} & { 0.1\%} & { 39.2\%} \\
\midrule
 & { \textbf{Factuality}} & { 4088} & { 21} & { 4083} & { 0.3\%} & \textbf{ 50.0\%} \\
 & { \textbf{Abstractiveness}} & { 3708} & { 31} & { 4453} & { 0.4\%} & \textbf{ 54.6\%} \\
 & { \textbf{Readability}} & { 3154} & { 2349} & { 2689} & { 28.7\%} & { 46.0\%} \\
\multirow{-4}{*}{\begin{tabular}[c]{@{}c@{}}Google\\ Local\\ Review\end{tabular}} & { \textbf{Overall}} & { 3750} & { 0} & { 4442} & { 0.0\%} & \textbf{ 54.2\%} \\
\bottomrule
\end{tabular}
\caption{Automated Evaluation Results on Comparing RLAIF v.s. RLPF. Note that the total number of examples may fluctuate as we exclude instances where either response doesn't adhere to the guidelines.}
\label{tab:autoeval_rlaif}
\end{table*}


\section{Direct RLAIF Baseline}
\label{app:rlaif}

Following the previous work \cite{harrison_rlaif_2024}, we used the following prompt(in Table~\ref{rlaif_prompt}) to compute reward scores for generated summaries during RLAIF training.
\begin{tcolorbox}
{\fontfamily{cmss}\selectfont
A good summary is a shorter piece of text that has the essence of the original. It tries to accomplish the same purpose and conveys the key information from the original input text. Below we define four evaluation axes for summary quality: coherence, accuracy, coverage, and overall quality.\\
~\\
Coherence: This axis answers the question “how coherent is the summary on its own?” A summary is coherent if it's easy to understand when read on its own and free of English errors. A summary is not coherent if it's difficult to understand what the summary is trying to say. Generally, it's more important that the summary is understandable than it being free of grammar errors.

Accuracy: This axis answers the question “does the factual information in the summary accurately match the input user history?” A summary is accurate if it doesn't say things that aren't in the input user hisory and generally is not misleading.

Coverage: This axis answers the question “how well does the summary cover the important information in the input user history?” A summary has good coverage if it mentions the main information from the input user history text that's important to understand the situation described in the user history. A summary has poor coverage if someone reading only the summary would be missing several important pieces of information about the situation in the input text.

Overall quality: This axis answers the question “how good is the summary overall at representing the input user history?” This can encompass all of the above axes of quality, as well as others you feel are important. If it's hard to find ways to make the summary better, the overall quality is good. If there are lots of different ways the summary can be made better, the overall quality is bad.

~\\
You are an expert user summary rater. Given a USER\_HISTORY\_TEXT as user past activity history in text and a SUMMARY that summarizes users' preference based on the past activity history, your role is to provide a SCORE from 1 to 10 that rates the quality of the SUMMARY given the USER\_HISTORY\_TEXT, with 1 being awful and 10 being a perfect SUMMARY.\\
~\\
USER\_HISTORY\_TEXT: \{input\_activities\} \\
SUMMARY: \{summary\} \\
SCORE:
}
\end{tcolorbox}
\noindent\begin{minipage}{1.0\textwidth}
\captionof{table}{Prompt Used by RLAIF Baseline.}\label{rlaif_prompt}
\end{minipage}

\section{Dataset Details}
\label{app:dataset}

\begin{itemize}
    \item \textbf{Amazon Review}~\cite{amazon_data_2016} encompasses a vast collection of product reviews from Amazon, spanning various categories, offering insights into customer preferences. 
    We used domain \textit{Books} for training and additional our domains for dataset transfer evaluation.
    \item \textbf{Google Local Review}~\cite{yan2022personalized} contains user reviews and ratings for local businesses across different locations, reflecting user experiences and sentiments towards local services. We chose \textit{New York} state in our experiments.
    \item \textbf{MovieLens 2015}~\cite{Harper2015MovieLens} commonly known as MovieLens20M, a widely used benchmark dataset for personalization and recommender systems. 
    \item \textbf{MovieLens 2003}~\cite{Harper2015MovieLens} also known as MovieLens1M.
    Unlike MovieLens 2015, this dataset has different users, movies, and collection times, plus user demographic data.
\end{itemize}
See Table \ref{tab:dataset} for detailed dataset statistics.

\begin{table*}[h]
\begin{tabular}{c|c|c|r|r|r|r|c}
\toprule
\textbf{Dataset} & \textbf{Abbreviation} & \textbf{Usage} & \multicolumn{1}{c|}{\begin{tabular}[c]{@{}c@{}}\textbf{\# Total}\\ \textbf{Items}\end{tabular}} & \multicolumn{1}{c|}{\begin{tabular}[c]{@{}c@{}}\textbf{\# Users}\\ \textbf{in Train}\end{tabular}} & \multicolumn{1}{c|}{\begin{tabular}[c]{@{}c@{}}\textbf{\# Train}\\ \textbf{Examples}\end{tabular}} & \multicolumn{1}{c|}{\begin{tabular}[c]{@{}c@{}}\textbf{\# Test}\\ \textbf{Examples}\end{tabular}} & \multicolumn{1}{c}{\begin{tabular}[c]{@{}c@{}}\textbf{Sequence}\\ \textbf{Length $N$}\end{tabular}} \\ \midrule
MovieLens 2015 & ML 2015 & Train \& Eval & 27,278 & 82,977 & 13,821,405 & 8,192 & 50 \\ \midrule
MovieLens 2003 & ML 2003 & Eval Only & 3,882 & 4,247 & 716,003 & 4,247 & 50 \\ \midrule
Google Local Review & Google Local & Train \& Eval & 270,721 & 80,080 & 3,387,375 & 8,192 & 50 \\ \midrule
\begin{tabular}[c]{@{}c@{}}Amazon Review\\ - Books\end{tabular} & Amazon Books & Train \& Eval & 2,935,525 & 64,191 & 4,936,920 & 8,192 & 50 \\ \midrule
\begin{tabular}[c]{@{}c@{}}Amazon Review\\ - Movies and TV\end{tabular} & Amazon Movies & Eval Only & 203,970 & 5,516 & 356,329 & 5,516 & 50 \\ \midrule
\begin{tabular}[c]{@{}c@{}}Amazon Review\\ - CDs and Vinyl\end{tabular} & Amazon CDs & Eval Only & 544,442 & 3,558 & 273,231 & 3,558 & 50 \\ \midrule
\begin{tabular}[c]{@{}c@{}}Amazon Review \\ - Toys and Games\end{tabular} & Amazon Toys & Eval Only & 634,414 & 1,242 & 41,741 & 1,242 & 50 \\ \midrule
\begin{tabular}[c]{@{}c@{}}Amazon Review\\ - Video Games\end{tabular} & Amazon Games & Eval Only & 84,893 & 447 & 16,589 & 447 & 50 \\ 
\bottomrule
\end{tabular}
\caption{Details of Datasets Used in Experiments.}
\label{tab:dataset}
\vskip -0.1in
\end{table*}

\section{Training Hyperparameters}
\label{app:hyperparameter}
During training, we set the policy temperature to 1.0 for Gemini 1.0 Nano-2 and 0.9 for PaLM-2 XS, while maintaining a prediction temperature of 0.1 during reward computation. We utilized the Adafactor optimizer \cite{shazeer2018adafactor} with a learning rate of 1e-6 and 100 warm-up steps, during which only the value network was updated while the policy network remained frozen. The weight $w$ that controls the balance between prediction feedback reward and length reward was set to 1 and the KL divergence coefficient $\alpha$ was set to 0.01, and the length reward parameters were configured as $\mathcal{C}$ = 1e-4 and $\beta$ = 3e-3. The target length $\mathcal{L}$ in the length reward was set to 358, 242, and 268 for Amazon Books, Google Local Review, and MovieLens 2015, respectively, to match the Zero-Shot sumamries. The input lengths for the policy models were 1600, 1280, and 1024 for Amazon Books, Google Local Review, and MovieLens 2015, respectively, with a consistent output length of 512 across all datasets. Training was conducted for 15,000 steps for each dataset, and all reported evaluation results are based on the final checkpoint.

\section{Statistical Significance}
We ran a paired t-test in Table \ref{tab:t-test} comparing the future activity prediction performance of RLPF and the Gemini 1.0 Nano-2 Zero-Shot model. The results demonstrate that the improvements achieved by RLPF are statistically significant.
\begin{table}[h]
\centering
\begin{tabular}{lrr}
\toprule
\multicolumn{1}{c}{\textbf{Dataset}} & \multicolumn{1}{c}{{ \textbf{t-statistic}}} & \multicolumn{1}{c}{\textbf{p-value}} \\
\midrule
Amazon Books & { -9.4155} & { 6.01e-21} \\
Google Local & { -10.0829} & { 9.05e-24} \\
ML 2015 & { -15.3415} & { 2.15e-52} \\
\bottomrule
\end{tabular}
\caption{Paired t-test results comparing Gemini 1.0 Nano-2 Zero-Shot summary and RLPF summary in future activity prediction.}
\label{tab:t-test}
\end{table}

\section{Task Selection for Reward Computation}
\label{app:task_selection}
Even though RLPF offers the flexibility to leverage any task and seamlessly accommodates the combination of rewards from multiple tasks, training separate models for each downstream task or using multiple tasks for training is not necessary. This underscores the convenience and efficiency of RLPF by eliminating the need for extensive, complex model training and overhead. Results shown in Table \ref{table:transfer} are all from unseen tasks/datasets, trained solely on the Future Activity Prediction (FAP) task. Of 19 unseen tasks, 16 showed improvements, demonstrating that the model is generalizable and transferable even when a single task is used for reward computation.

We provide the following guidelines on task selection for reward computation to enhance the applicability for various needs:
\begin{itemize}
    \item \textbf{Broad Contextual Use} If the summary is intended to support a wide range of downstream tasks, selecting a single task that captures comprehensive user context is sufficient for transferability and generalizability. We elaborate how to choose this task below.
    \item \textbf{Single Most Important Task} If a personalization system has one primary task, using that task for reward computation yields the best performance for that individual task.
    \item \textbf{Flexibility for Multitask Rewards}: Our design can also accommodate multitask rewards, which can be useful if a system requires optimized performance across a specific set of tasks.
\end{itemize}

\subsection{Reason for choosing FAP}
FAP was chosen in our experiments for reward computation due to its wide use in personalization systems, capturing user preferences and sequential behavior. We also tested using Favorite Genre Prediction (FGP) and Rating Prediction (RP) for reward computation as alternatives. See Table \ref{tab:reward_task} for relative changes compared with the baseline model. The results show that using a task for reward computation does yield best performance for that individual task. While FGP and RP supported transferability between each other, they did not enable effective transfer to FAP, showing that the chosen task should capture both user preferences and sequential patterns in order to achieve broad transferability.

\begin{table}[h]
\centering
\begin{tabular}{c|ccc}
\toprule
\multicolumn{1}{c|}{} & \multicolumn{3}{c}{Eval Task} \\
\cmidrule(lr){2-4}
Training Task & FAP & FGP & RP \\
\midrule
FAP & 18.46\% & 5.68\% & 3.11\% \\
FGP & -7.40\% & 24.55\% & 1.78\% \\
RP & -2.79\% & 7.36\% & 8.00\% \\
\bottomrule
\end{tabular}
\caption{Performance gain of using alternative tasks for reward computation. FAP: Future Activity Prediction. FGP: Favorite Genre Prediction. RP: Rating Prediction.}
\label{tab:reward_task}
\end{table}

\section{Grounded Evaluation Metrics for Readability}
\label{app:fres}
We computed the Flesch reading-ease score (FRES)~\cite{imperial2023} in Table \ref{tab:fres} , a common metric to evaluate readability, for raw activity data and summaries before and after applying RLPF. The results demonstrated that user summaries exhibit significantly better readability than raw activity data. However, there are mixed results between pre- and post- RLPF training across three datasets. 

While Flesch-Kincaid provides a quick and established method for evaluating readability, it may not be the best way to measure the readability of user summaries. User summaries are meant to capture nuanced behavior patterns and contexts that Flesch-Kincaid scores cannot measure. Additionally, the formula's assumption that shorter sentences and simpler words are inherently more readable does not always hold true for user summaries. This limitation is the main reason that we get a lower Post-RLPF summary readability score on Google Review dataset, and the LLM's readability score aligns better with human judgement \cite{trott-riviere-2024-measuring}.

\begin{table}[h]
\centering
\begin{tabular}{lccc}
\toprule
\multicolumn{1}{c}{\textbf{Dataset}} & \multicolumn{1}{c}{\textbf{\begin{tabular}[c]{@{}c@{}}Raw User \\ Activity Data\end{tabular}}} & \multicolumn{1}{c}{\textbf{\begin{tabular}[c]{@{}c@{}}Pre-RLPF\\ Summary\end{tabular}}} & \multicolumn{1}{c}{\textbf{\begin{tabular}[c]{@{}c@{}}Post-RLPF\\ Summary\end{tabular}}} \\
\midrule
ML2015 & -64 & 60 & 64 \\
Amazon Review & -114 & 12 & 20 \\
Google Local & -283 & 47 & 23 \\
\bottomrule
\end{tabular}
\caption{Flesch reading-ease score (FRES) for evaluation readability of user raw activity data and user summary before and after RLPF training.}
\label{tab:fres}
\end{table}

\newpage
\section{Detailed Results}
\label{app:results}

See Table \ref{tab:main_result_numbers} for more detailed results for future activity prediction and favorite genre/category predction. See Table \ref{tab:token_length} for input token length comparison between summariesa and full user context.

\begin{table*}[h]
\begin{tabular}{lccccccc}
\toprule
\multicolumn{1}{c}{\multirow{2}{*}{\textbf{Prediction Input}}} & \multicolumn{2}{c}{\textbf{Amazon Books}} & \multicolumn{2}{c}{\textbf{Google Local Review}} & \multicolumn{2}{c}{\textbf{MovieLens 2015}} & \textbf{Overall} \\
\multicolumn{1}{c}{} & \makecell{Future\\Activity} & \textit{\makecell{Fav\\Category*}} & \makecell{Future\\Activity} & \textit{\makecell{Fav\\Category*}} & \makecell{Future\\Activity} & \textit{\makecell{Fav\\Category*}} & Average \\
\midrule
First $X$ Activities & 0.497 & 0.586 & 0.451 & 0.467 & 0.640 & 0.763 & 0.567 \\
Random $X$ Activities & 0.489 & 0.590 & 0.451 & 0.473 & 0.668 & 0.798 & 0.578 \\
Last $X$ Activities & 0.582 & 0.587 & 0.492 & 0.472 & \textbf{0.738} & 0.753 & 0.604 \\
All $N$ Activities & 0.530 & 0.596 & 0.507 & 0.494 & 0.733 & \textbf{0.910} & 0.628 \\
\midrule
Gemini 1.0 Nano-2 0-Shot & 0.664 & 0.594 & 0.483 & 0.487 & 0.607 & 0.774 & 0.601 \\
Gemini 1.0 Pro 0-Shot & 0.647 & 0.585 & 0.504 & 0.488 & 0.595 & 0.843 & 0.610 \\
Gemini 1.0 Pro Few-Shot & 0.680 & 0.609 & 0.543 & 0.520 & 0.654 & 0.805 & 0.635\\
Gemini 1.0 Nano-2 Crafted Prompt & 0.684 & 0.541 & 0.523 & 0.419 & 0.653 & 0.899 & 0.620 \\
Gemini 1.0 Nano-2 RLAIF & 0.654 & \textbf{0.613} & 0.458 & 0.513 & 0.519 & 0.776 & 0.589 \\
\midrule
Gemini 1.0 Nano-2 RLPF (ours) & \textbf{0.709} & 0.605 & \textbf{0.560} & \textbf{0.559} & 0.719 & 0.818 & \textbf{0.662} \\
\textit{Improv over Nano-2 0-shot} & \textit{6.78\%} & \textit{1.85\%} & \textit{15.93\%} & \textit{14.78\%} & \textit{18.46\%} & \textit{5.68\%} & \textit{10.00\%} \\
\textit{Improv over All N Activities} & \textit{33.77\%} & \textit{1.51\%} & \textit{10.44\%} & \textit{13.15\%} & \textit{-1.91\%} & \textit{-10.11\%} & \textit{5.30\%} \\
\bottomrule
\end{tabular}
\caption{Evaluation Results on \textit{Future Activity Prediction} and \textit{Favorite Category Prediction} Tasks between Different Approaches. * denoted as unseen tasks from training. $X$ = 16, 10, 13 for Amazon Books, Google Local Review, MovieLens 2015 respectively to keep the token length comparable with RLPF summaries.}
\label{tab:main_result_numbers}
\end{table*}

\begin{table*}[h]
\centering
\begin{tabular}{lcccc}
\toprule
 & \multicolumn{1}{c}{\textbf{Amazon Books}} & \multicolumn{1}{c}{\textbf{Google Local}} & \multicolumn{1}{c}{\textbf{ML 2015}} & \multicolumn{1}{c}{\textbf{Average}} \\
\midrule
All N Activities & 812 & 642 & 791 & 748 \\
Gemini 1.0 Nano-2 0-Shot & 349 & 260 & 272 & 293 \\
Gemini 1.0 Nano-2 RLPF (ours) & \textbf{267} & \textbf{133} & \textbf{198} & \textbf{200} \\
\midrule
\begin{tabular}[c]{@{}l@{}}RLPF Compression Ratio\\ Over All N Activities\end{tabular} & 67.0\% & 79.3\% & 74.9\% & 73.8\% \\
\bottomrule
\end{tabular}
\caption{Average Token Length of Different Approaches.}
\label{tab:token_length}
\end{table*}

\section{Task Definition Details}
\label{app:task}

\begin{table*}[h!]
\begin{tabular}{m{2.2cm}| m{2.5cm}| m{1.5cm}| m{2cm}|m{7cm}}
\toprule
\multicolumn{1}{c}{\textbf{Task Name}} & \multicolumn{1}{c}{\textbf{Dataset}} & \multicolumn{1}{c}{\textbf{Metric}} & \multicolumn{1}{c}{\textbf{Abbreviation}} & \multicolumn{1}{c}{\textbf{Detailed Description}} \\
\midrule
\begin{tabular}[c]{@{}l@{}}Future Activity\\ Prediction\end{tabular} & \begin{tabular}[c]{@{}l@{}}Amazon Books\\ Amazon Movies\\ Amazon CDs\\ Amazon Toys\\ Amazon Games\\ Google Local\\ ML 2015\\ ML 2003\end{tabular} & Accuracy & Future Act & Predict the item name (e.g., movie, product, or place) the user will interact with in the future, presented in a multiple-choice format with three randomly selected negative options. This task serves as the sole task used in both training and evaluation. \\
\midrule
\begin{tabular}[c]{@{}l@{}}Favorite Genre\\ or Category\\ Prediction\end{tabular} & \begin{tabular}[c]{@{}l@{}}Amazon Books\\ Google Local\\ ML 2015\\ ML 2003\end{tabular} & Recall@3 & \begin{tabular}[c]{@{}l@{}}Fav Genre or\\ Fav Category\end{tabular} & Predict the user's favorite genre or category. The label is determined as the most frequent genre/category appearing in the entire user's history. For MovieLens, the model predicts three out of 19 genres. For Amazon Books and Google Local Review, due to their larger category vocabularies, we limit the model's choices to the user's top 10 most frequent categories, again prompting for three predictions. Importantly, the genre/category information for each item is not included in the user context provided for summary generation. This task evaluates the summarization model's ability to infer high-level user insights based on real-world knowledge. \\
\midrule
\begin{tabular}[c]{@{}l@{}}Rating\\ Prediction\end{tabular} & \begin{tabular}[c]{@{}l@{}}Amazon Books\\ Google Local\\ ML 2015\\ ML 2003\end{tabular} & Accuracy & Rating & Predict the rating a user would assign to a specified item. In our experiments, we specified the item as the target activity in Future Activity Prediction. \\
\midrule
\begin{tabular}[c]{@{}l@{}}Common\\ Activity \\ Retrieval\end{tabular} & Google Local & Recall@3 & Common City & Predict the city the user has visited most frequently on Google Local Review dataset. This task assesses whether a user summary can furnish sufficient information for user context retrieval, a crucial functionality for agent systems. \\
\midrule
\begin{tabular}[c]{@{}l@{}}Review\\ Generation\end{tabular} & Amazon Books & ROUGE-Lsum & Review Gen & Given two exemplar reviews written by the user for other items, generate a review for a specified item that aligns with the user's writing style. \\
\midrule
\begin{tabular}[c]{@{}l@{}}User Age\\ Prediction\end{tabular} & ML 2003 & Accuracy & User Age & Predict the user's age range. \\
\midrule
\begin{tabular}[c]{@{}l@{}}User Gender\\ Prediction\end{tabular} & ML 2003 & Accuracy & User Gender & Predict the user's gender. \\
\midrule
\begin{tabular}[c]{@{}l@{}}User Occupancy\\ Prediction\end{tabular} & ML 2003 & Recall@3 & \begin{tabular}[c]{@{}l@{}}User\\ Occupancy\end{tabular} & Predict the user's occupancy. \\
\bottomrule
\end{tabular}
\caption{Prediction Task Information and Definitions.}
\label{tab:task_definition}
\end{table*}

Detailed definitions and information about tasks in experiments are listed in Table \ref{tab:task_definition}. Prompts that are used for predictions are listed in Table \ref{tab:prompt_future_act_prediction} in Appendix \ref{app:prompt_unseen_tasks}.


\section{Example Summaries}
\label{app:example_summary}
\begin{figure*}
\centering
    \includegraphics[width=0.99 \textwidth]{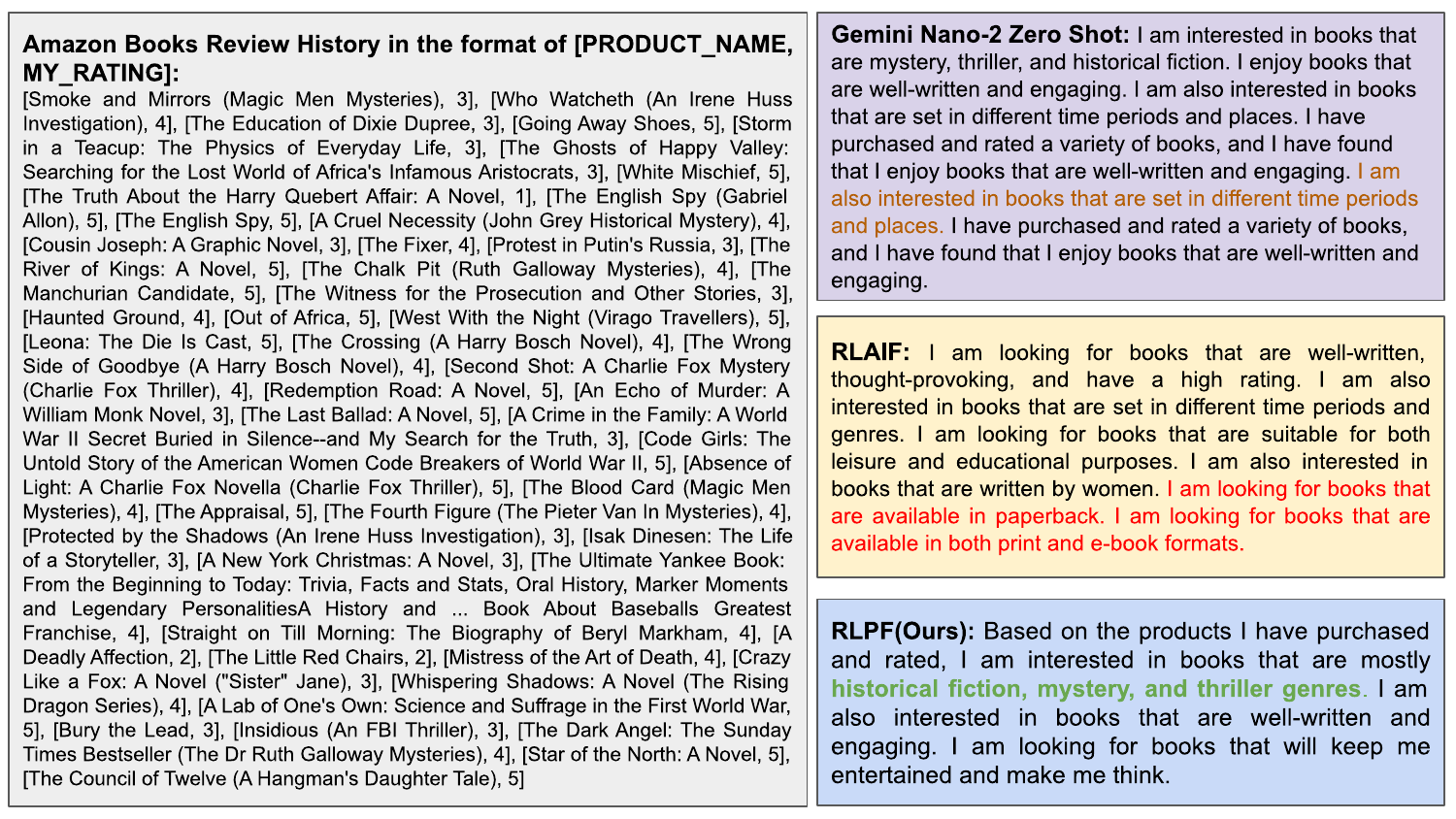}
    \vspace{-0.08in}
    \caption{Example Summary Comparison on \textbf{Amazon Books}. There are duplicate parts (\textcolor{orange}{orange}) in Zero-Shot summary, and hallucinations in RLAIF summary (\textcolor{red}{red}), while summary generated by RLPF adheres to the facts highlighted in \textcolor{green}{\textbf{green}}. Activity data statistics (out of 50 book review activities): Fiction: 34, Thrillers: 12, Written by woman: 20.}
    \vskip -0.1in
    \label{fig:example_sum_amazon_1}
\end{figure*}

\begin{figure*}
\centering
    \includegraphics[width=0.99 \textwidth]{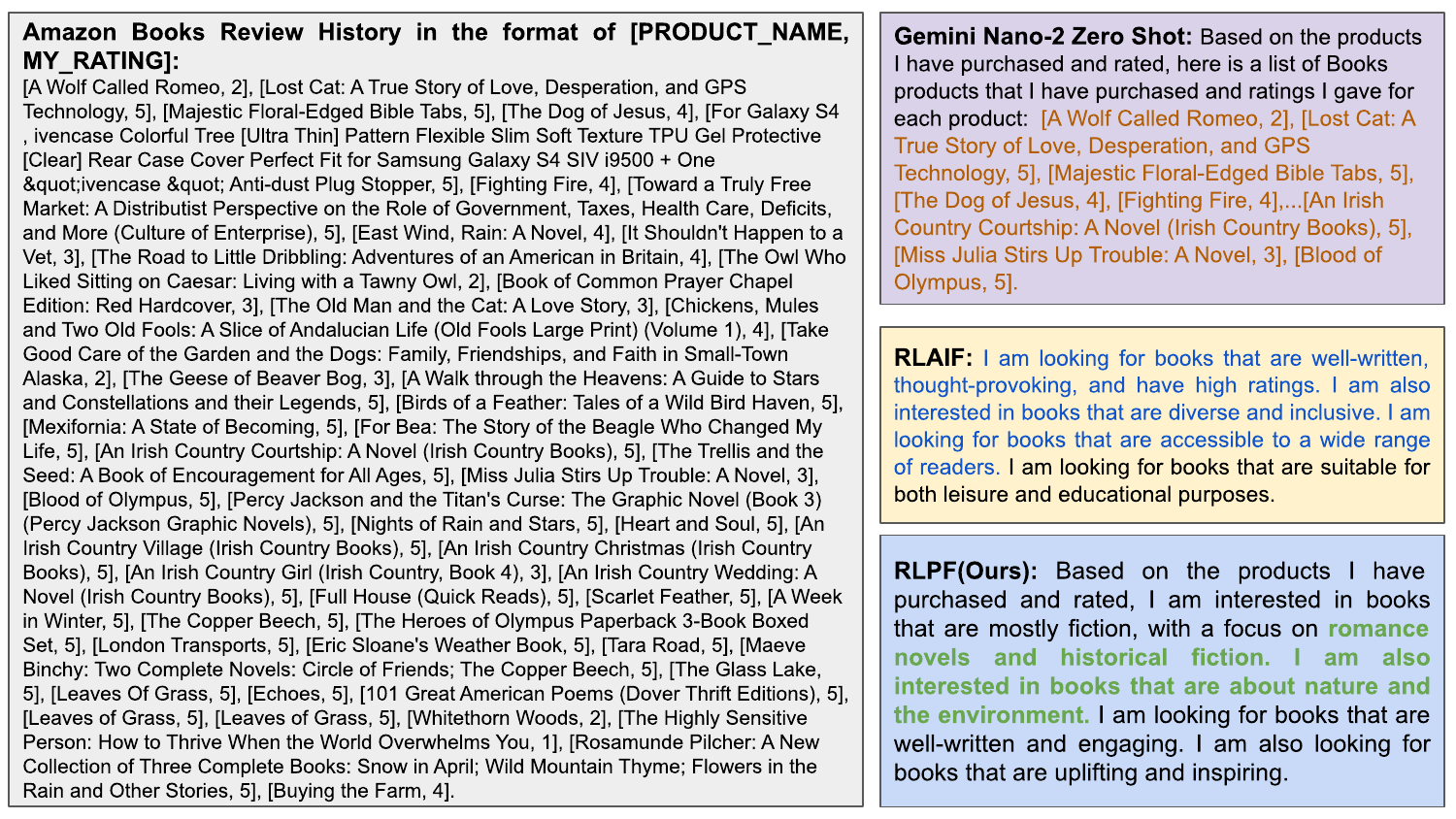}
    \vspace{-0.08in}
    \caption{Example Summary Comparison on \textbf{Amazon Books}. There are duplicate parts (\textcolor{orange}{orange}) in Zero-Shot summary, and too general summary in RLAIF (\textcolor{blue}{blue}), while summary generated by RLPF adheres to the facts highlighted in \textcolor{green}{\textbf{green}}. Activity data statistics (out of 50 book review activities): Fiction: 25, Romance Novels: 11, Historical: 4, Books about Nature or Environment: 4.}
    \vskip -0.1in
    \label{fig:example_sum_amazon_2}
\end{figure*}

\begin{figure*}
\centering
    \includegraphics[width=0.99 \textwidth]{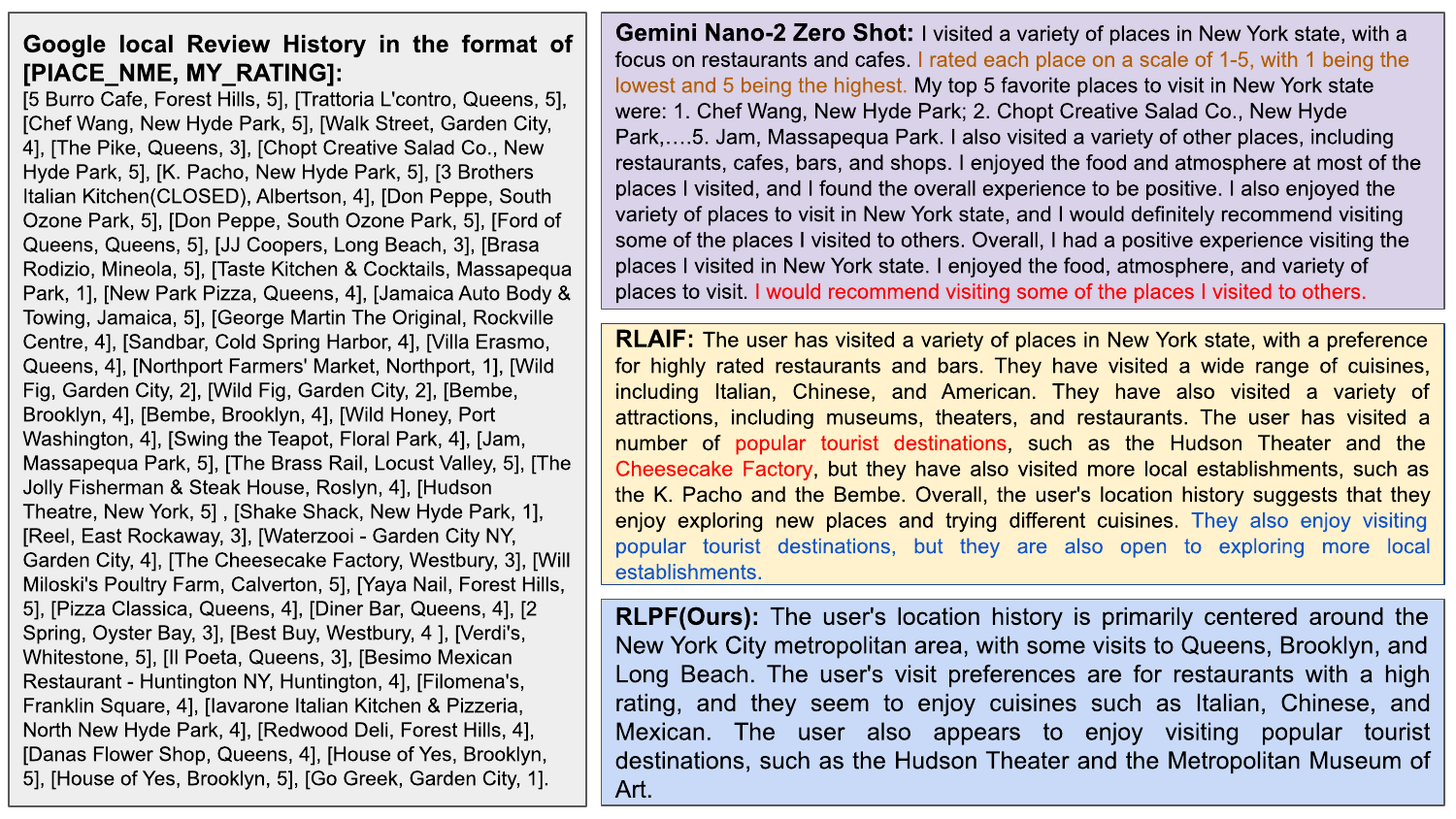}
    \vspace{-0.08in}
    \caption{Example Summary Comparison on \textbf{Google Local Review}. There are duplicate parts (\textcolor{orange}{orange}) and hallucination in both Zero-Shot summary. RLAIF's summary also has the issue of being too general summary (\textcolor{blue}{blue}). Activity data statistics (out of 50 place review activities): Italian: 7, Chinese: 2, Mexican: 2, Popular Tourist Destinations: 3.}
    \vskip -0.1in
    \label{fig:example_sum_google_1}
\end{figure*}

\begin{figure*}
\centering
    \includegraphics[width=0.99 \textwidth]{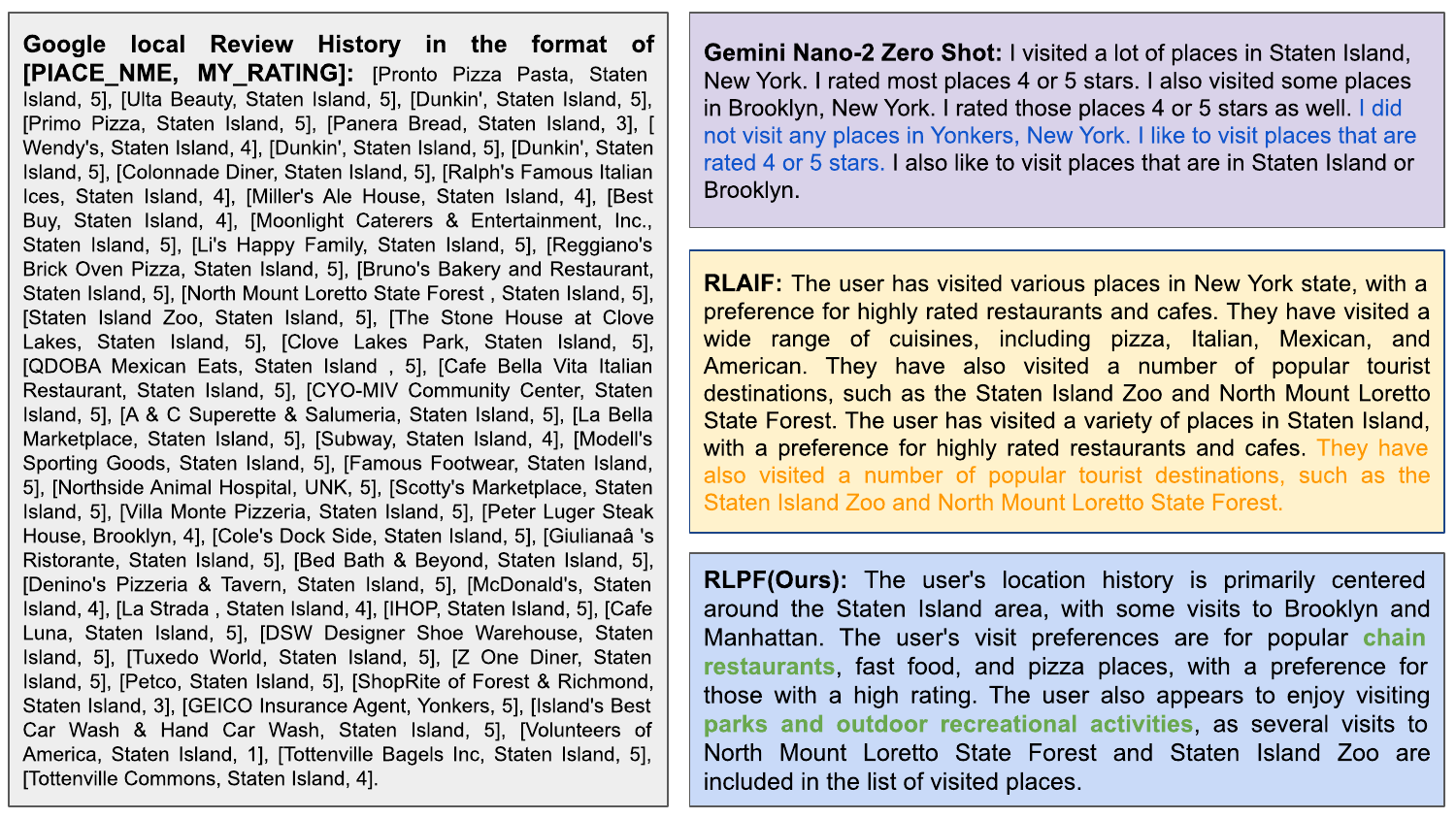}
    \vspace{-0.08in}
    \caption{Example Summary Comparison on \textbf{Google Local Review}. Zero-Shot summary is too general (\textcolor{blue}{blue}) while RLAIF's summary contains of duplication (\textcolor{orange}{orange}). Activity data statistics (out of 50 place review activities): Italian Restaurants: 9, Mexico: 1, American: 3, Restaurants chain: 7.}
    \vskip -0.1in
    \label{fig:example_sum_google_2}
\end{figure*}

\begin{figure*}
\centering
    \includegraphics[width=0.99 \textwidth]{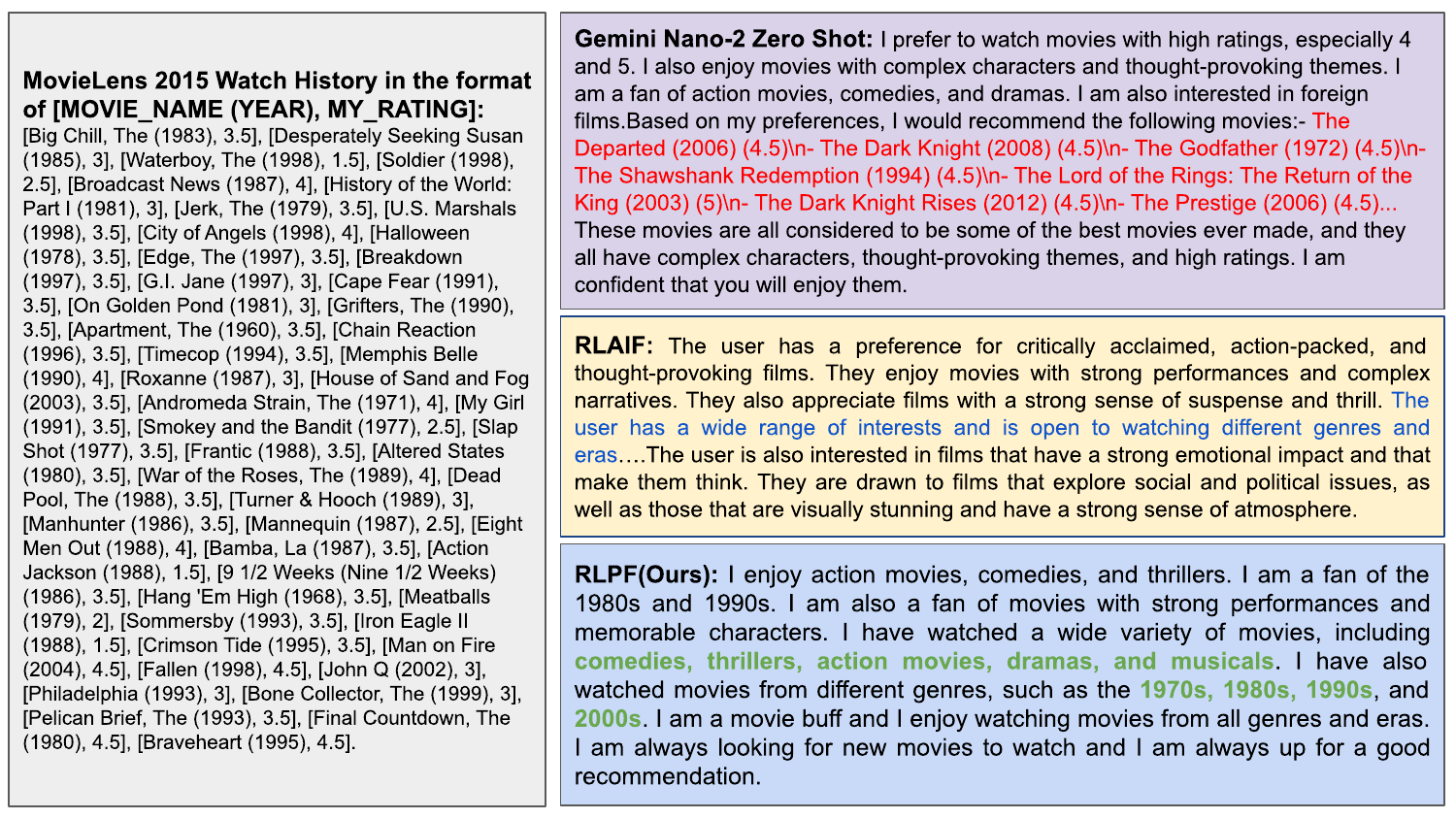}
    \vspace{-0.08in}
    \caption{Example Summary Comparison on \textbf{MovieLens 2015}. There are hallucination, which doesn't exist in the input activities (\textcolor{red}{red}) in Zero-Shot summary. RLAIF's summary also has the issue of being too general summary (\textcolor{blue}{blue}), while summary generated by RLPF adheres to the facts (\textcolor{green}{\textbf{green}}). Activity data statistics (out of 50 movie watch activities): Drama: 19, Action: 11, Comedy: 10, Thriller: 9, Science Fiction: 5, Romance: 5, Crime: 5, Sports: 4, Historical: 4, Horror: 2.}
    \vskip -0.1in
    \label{fig:example_mlm_1}
\end{figure*}

\begin{figure*}
\centering
    \includegraphics[width=0.99 \textwidth]{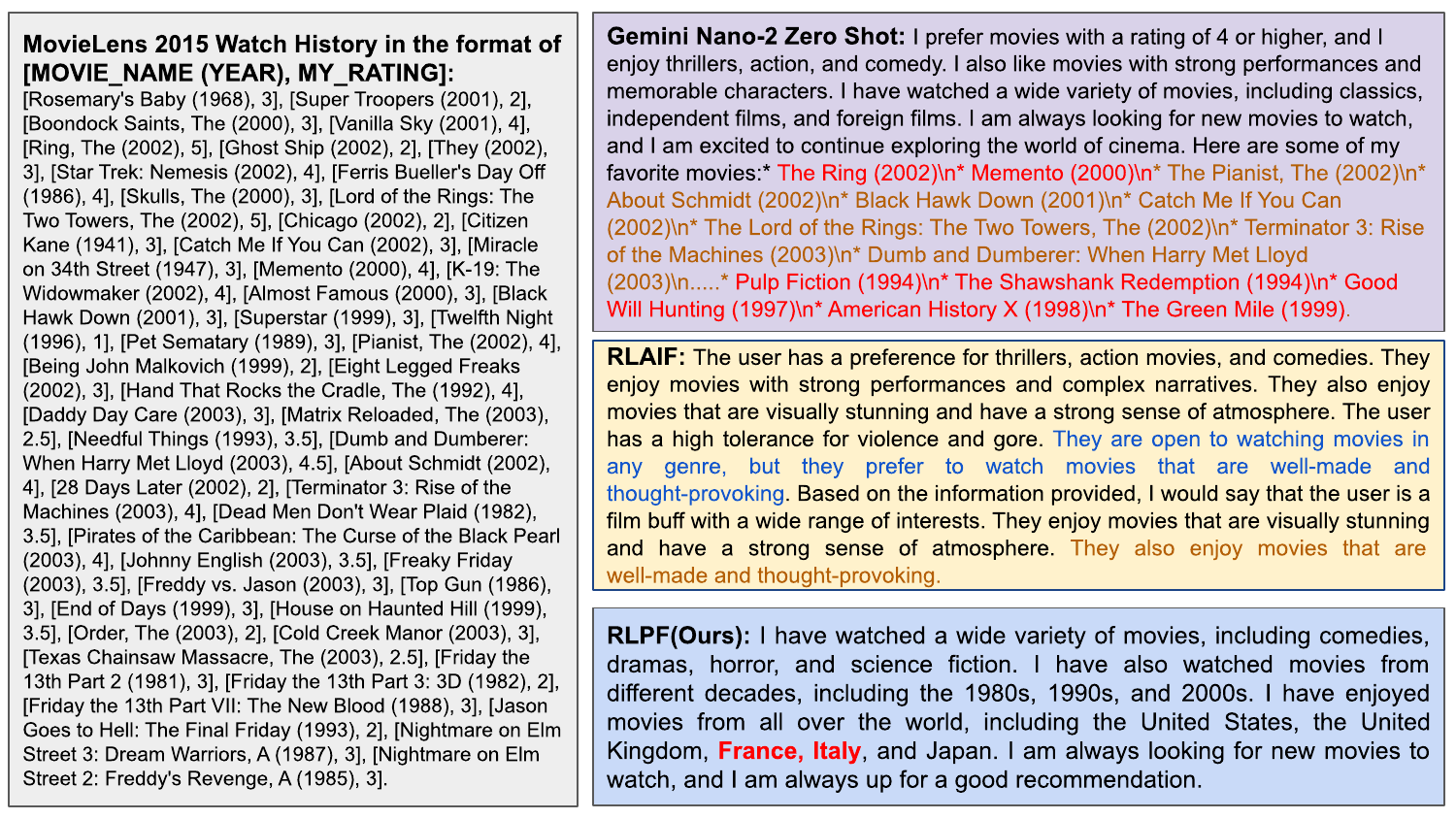}
    \vspace{-0.08in}
    \caption{Example Summary Comparison on \textbf{MovieLens 2015}. There are hallucination, which doesn't exist in the input activities (\textcolor{red}{red}) as well as simple repetition of the input activities (\textcolor{orange}{orange}) in Zero-Shot summary. RLAIF's summary also suffers the issue of being too general (\textcolor{blue}{blue}) or repetition (\textcolor{orange}{orange}). We also observe that summary generated by RLPF has some hallucination (\textcolor{red}{\textbf{red}}). Activity data statistics (out of 50 movie watch activities): Comedies: 5, Dramas: 7, Horror: 13, Science Fiction: 3. United Kingdom Movie: 1, France Movie: 0, Italy Movie: 0, Japan Movie: 1.}
    \vskip -0.1in
    \label{fig:example_mlm_2}
\end{figure*}

Example summaries are displayed in Figure \ref{fig:example_sum_amazon_1},\ref{fig:example_sum_amazon_2},\ref{fig:example_sum_google_1},\ref{fig:example_sum_google_2},\ref{fig:example_mlm_1},\ref{fig:example_mlm_2}. 

\onecolumn
\newpage

\onecolumn
\newpage
\section{Prompts}
\label{app:prompt}

\begin{table*}[h]
\small
\begin{tabular}{m{2cm}|b{4cm}|b{10cm}}
\toprule
\multicolumn{1}{c}{\textbf{Dataset}} & \multicolumn{1}{c}{\textbf{Zero-Shot/RLPF Prompt}} & \multicolumn{1}{c}{\textbf{Crafted Prompt}} \\
\midrule
\begin{tabular}[c]{m{2cm}}MovieLens\\ 2015\\ MovieLens\\ 2003\end{tabular} & \begin{tabular}[c]{b{3.5cm}}Here is a list of movies that I have watched in the format of {[}MOVIE\_NAME (YEAR), MY\_RATING{]}, where RATING=1 is the lowest and RATING=5 is the highest:\\ \{input\_activities\}\\ Please summarize my movie watch preference within 200 words.\end{tabular} & \textit{\textbf{\begin{tabular}[c]{p{10cm}}Summarize my movie watch history from three aspects.\\ First, list three of my favorite movie genres.\\ Second, summarize the time periods of movies I like, e.g. 1980s, 1990s etc.\\ Third, provide five movies belonging to my favorite genres.\\ \\ For example:\\     Favorite movie genres are \textless{}GENRE1\textgreater{}, \textless{}GENRE2\textgreater{}, \textless{}GENRE1\textgreater\\     Time periods of liked movies are \textless{}TIME1\textgreater{}, \textless{}TIME2\textgreater\\     Example movies are \textless{}MOVIE1\textgreater{}, \textless{}MOVIE2\textgreater{}, \textless{}MOVIE3\textgreater{}, \textless{}MOVIE4\textgreater{}, \textless{}MOVIE5\textgreater\\ 
~\\
 {\normalfont Here is a list of movies that I have watched in the format of {[}MOVIE\_NAME (YEAR), MY\_RATING{]}, where RATING=1 is the lowest and RATING=5 is the highest:}\\ {\normalfont \{input\_activities\}}\\ {\normalfont Please summarize my movie watch preference within 200 words.}\end{tabular}}} \\
\midrule
\begin{tabular}[c]{m{2cm}}Amazon\\ Review\end{tabular} & \begin{tabular}[c]{b{3.5cm}}Here is a list of \{domain\} products that I have purchased and ratings I gave for each product in the format of {[}PRODUCT\_NAME, MY\_RATING{]}, where MY\_RATING=1 is the lowest and MY\_RATING=5 is the highest:\\ \{input\_activities\}\\ Please summarize my interest about \{domain\} within 200 words.\end{tabular} & \textit{\textbf{\begin{tabular}[c]{b{10cm}}Summarize my interest about \{domain\} products from two aspects.\\ First, list three of my favorite categories of \{domain\}.\\ Second, provide five products belonging to my favorite categories of \{domain\}.\\ \\ For example:\\     Favorite categories of \{domain\} are \textless{}CATEGORY1\textgreater{}, \textless{}CATEGORY2\textgreater{}, \textless{}CATEGORY3\textgreater\\     Example products are \textless{}PRODUCT1\textgreater{}, \textless{}PRODUCT2\textgreater{}, \textless{}PRODUCT3\textgreater{}, \textless{}PRODUCT4\textgreater{}, \textless{}PRODUCT5\textgreater\\ 
~\\
{\normalfont Here is a list of \{domain\} products that I have purchased and ratings I gave for each product in the format of {[}PRODUCT\_NAME, MY\_RATING{]}, where MY\_RATING=1 is the lowest and MY\_RATING=5 is the highest:}\\ {\normalfont\{input\_activities\}}\\ {\normalfont Please summarize my interest about \{domain\} within 200 words.}\end{tabular}}} \\
\midrule
\begin{tabular}[c]{m{2cm}}Google\\ Local\\ Review\end{tabular} & \begin{tabular}[c]{b{3.5cm}}Here is a list of places that I have visited in \{state\} state and ratings I gave for each place in the format of {[}PLACE\_NAME, CITY\_NAME, MY\_RATING{]}, where MY\_RATING=1 is the lowest and MY\_RATING=5 is the highest: \\ \{input\_activities\}\\ Please summarize my location history pattern and place visit preference within 200 words.\end{tabular} & \textit{\textbf{\begin{tabular}[c]{b{10cm}}Summarize my location history pattern and place visit preference from three aspects.\\ First, list three of my favorite categories for places to visit in \{state\} state.\\ Second, list three common cities I visited in \{state\} state.\\ Third, provide five places belonging to my favorite categories for places to visit.\\ \\ For example:\\     Favorite categories for places to visit are \textless{}CATEGORY1\textgreater{}, \textless{}CATEGORY2\textgreater{}, \textless{}CATEGORY3\textgreater\\     Common cities I visited are \textless{}CITY1\textgreater{}, \textless{}CITY2\textgreater{}, \textless{}CITY3\textgreater\\     Example places are \textless{}PLACE1\textgreater{}, \textless{}PLACE2\textgreater{}, \textless{}PLACE3\textgreater{}, \textless{}PLACE4\textgreater{}, \textless{}PLACE5\textgreater\\ 
~\\
{\normalfont Here is a list of places that I have visited in New York state and ratings I gave for each place in the format of {[}PLACE\_NAME, CITY\_NAME, MY\_RATING{]}, where MY\_RATING=1 is the lowest and MY\_RATING=5 is the highest:} \\ \normalfont{\{input\_activities\}}\\ \normalfont{Please summarize my location history pattern and place visit preference within 200 words.}\end{tabular}}}\\
\bottomrule
\end{tabular}
\caption{Different summarization prompts. Zero-Shot/RLPF Prompt: This prompt is utilized for both training the summarizer model and evaluating the "Gemini 1.0 Nano-2 Zero-Shot" baseline. Crafted Prompt: Employed for baseline evaluation with "Gemini 1.0 Nano-2 with Crafted Prompts." The highlighted text within these prompts represents additional instructions tailored to specific downstream task predictions. For Amazon Review, \{\textit{domain}\} is replaced with Books, Movies and TV, CDs and Vinyl, Toys and Games, Video Games etc. \{\textit{domain}\} is replaced with New York. }
\label{tab: summarization_prompts}
\end{table*}

\begin{table*}[h]
\centering
\begin{tabular}{ll}
\toprule
\multicolumn{1}{c}{\textbf{Dataset}} & \multicolumn{1}{c}{\textbf{Future Activity Prediction Prompt}}\\
\midrule
\begin{tabular}[c]{@{}l@{}}MovieLens\\ 2015\end{tabular} & \begin{tabular}[c]{@{}l@{}}Here is a summary of my movie watch preference (between \textless{}summary\textgreater and \textless{}/summary\textgreater{}).\\ \textless{}summary\textgreater{}\{summary\}\textless{}/summary\textgreater{}.\\ Base on this information, which movie will I watch next? Choose one from the following options:\\ \{choices\}\\ Structure the answer with '{[}Index of your choice(A/B/C/D){]}'.\\ Answer:\end{tabular}\\
\midrule
\begin{tabular}[c]{@{}l@{}}Amazon\\ Review\end{tabular} & \begin{tabular}[c]{@{}l@{}}Here is a summary of my product purchase preference (between \textless{}summary\textgreater and \textless{}/summary\textgreater{}).\\ \textless{}summary\textgreater{}\{summary\}\textless{}/summary\textgreater{}.\\ Base on this information, which product will I purchase next? Choose one from the following options:\\ \{choices\}\\ Structure the answer with '{[}Index of your choice(A/B/C/D){]}'.\\ Answer:\end{tabular}\\
\midrule
\begin{tabular}[c]{@{}l@{}}Google\\ Local\\ Review\end{tabular} & \begin{tabular}[c]{@{}l@{}}Here is a summary of my place visit preference (between \textless{}summary\textgreater and \textless{}/summary\textgreater{}).\\ \textless{}summary\textgreater{}\{summary\}\textless{}/summary\textgreater{}.\\ Base on this information, which place will I visit next? Choose one from the following options:\\ \{choices\}\\ Structure the answer with '{[}Index of your choice(A/B/C/D){]}'.\\ Answer:\end{tabular}\\
\bottomrule
\end{tabular}
\caption{Prompt for \textit{Future Activity Prediction}, used in both RLPF training during reward computation and evaluation.}
\label{tab:prompt_future_act_prediction}
\end{table*}

\subsection{Prompt for Summary Generation and Future Activity Prediction}

See Table \ref{tab: summarization_prompts} and for full prompts we used to generate summaries during training and evaluation baselines, and Table \ref{tab:prompt_future_act_prediction} for future activity prediction prompt used during reward computation and evaluation. 

\subsection{Different Summarization and Prediction Prompts to Assess Prompt Robustness}
Different summarization and prediction prompts used in Figure ``Prompt Robustness'' are shown in Table \ref{tab:prompt_robustness}. In addition, all the prediction prompts for unseen tasks used in experiments are listed below.

\begin{table*}[]
\centering
\small
\begin{tabular}{m{2.5cm}|m{4cm}|m{9cm}}
\toprule
\multicolumn{1}{c}{\textbf{Usage}} & \multicolumn{1}{c}{\textbf{Prompt 1}} & \multicolumn{1}{c}{\textbf{Prompt 2}} \\
\midrule
\large{Summarization} &
\begin{tabular}[c]{m{3.5cm}}
Here is a list of movies that I have watched in the format of [MOVIE\_NAME (YEAR), MY\_RATING], where RATING=1 is the lowest and RATING=5 is the highest: \\ 
\{input\_activities\}  \\
Please summarize my movie watch preference within 200 words.
\end{tabular} &
\begin{tabular}[c]{m{8.5cm}} 
Here is a list of movies that I have watched in the format of [MOVIE\_NAME (YEAR), MY\_RATING], where RATING=1 is the lowest and RATING=5 is the highest: \\
\{input\_activities\} \\
Please summarize my movie watch preference within 200 words. \textbf{A good summary is a shorter piece of text that has the essence of users' preference based on their past movie watch history. It also conveys the key information about the person based on their past watch history. The summary should be coherent and easy to understand when read on its own. The factual information in the summary should accurately match the person's past watch history.}  \\
\textbf{Summary}: 
\end{tabular} \\
\midrule
\large{Prediction} & 
\begin{tabular}[c]{m{3.5cm}}
Here is a summary of my movie watch preference (between \textless{}summary\textgreater{} and \textless{}/summary\textgreater{}). \\
\textless{}summary\textgreater{}\{summary\}\textless{}\\/summary\textgreater{}. \\
Base on this information, which movie will I watch next? Choose one from the following options:
\{choices\} \\
\textbf{Let's think it step by step and structure the answer with the following format: Rationale: [Why do you choose this option] and Answer: [Your Choice(A/B/C/D)].}
\end{tabular} & 
\begin{tabular}[c]{m{8.5cm}}
Here is a summary of my movie watch preference (between \textless{}summary\textgreater{} and \textless{}/summary\textgreater{}).\\
\textless{}summary\textgreater{}\{summary\}\textless{}/summary\textgreater{}. \\
Base on this information, which movie will I watch next? Choose one from the following options:
\{choices\} \\
\textbf{Structure the answer with '[Index of your choice(A/B/C/D)]'.} \\
\textbf{Answer:} 
\end{tabular} \\
\bottomrule
\end{tabular}
\caption{Prompts used in Figure ``Prompt Robustness''. Top: Different summarization prompts during training. Bottom: Different prediction prompts during reward computation. The additional texts are highlighted.}
\label{tab:prompt_robustness}
\end{table*}

\subsection{Prompts for Unseen Prediction Tasks}
\label{app:prompt_unseen_tasks}

\subsubsection{MovieLens 2015 and MovieLens 2003\\}

All unseen prediction tasks for these two datasets share the same prompt preamble: 
\begin{tcolorbox}
\texttt{Here is a summary of my movie watch preference (between \textless{}summary\textgreater{} and \textless{}/summary\textgreater{}).\\
\textless{}summary\textgreater{}SUMMARY\textless{}/summary\textgreater{}
}
\end{tcolorbox}

For each prediction task, the prompt suffix is as following:
\begin{itemize}
    \item \textbf{Favorite Genre Prediction}:\\
    \begin{mdframed}[nobreak=true]
      \small {\texttt{Based on this information, what are my top 3 favorite movie genre? Choose three from the following genres: \\
      Drama, Comedy, Thriller, Action, Romance, Adventure, Crime, Sci-Fi, Horror, Fantasy, Mystery, Children, War, Animation, Musical, Documentary, Western, IMAX, Film-Noir. \\
      Structure the answer in the format of "GENRE1$\rightarrow$ GENRE2$\rightarrow$GENRE3". \\
      Answer:}}
      \end{mdframed}
      \vspace{0.3cm}
    \item \textbf{Rating Prediction}: \\
      \begin{mdframed}[nobreak=true]
      \small {\texttt{Based on this information, what is my rating on the following movie? \\
      Answer a score between 1 to 5 inclusive, using half-point(0.5) increments.
      Structure the answer with "[Your choice from (1.0, 1.5, 2.0, 2.5, 3.0, 3.5, 4.0, 4.5, 5.0)]."\\
      Answer:
      }}
      \end{mdframed}
      \vspace{0.3cm}
    \item \textbf{User Gender Prediction}: \\
      \begin{mdframed}[nobreak=true]
      \small {\texttt{Based on this information, what is my gender? Choose one from the following genders:\\
      Male, Female.\\
      Structure the answer with one gender only without extra words.\\
      Answer:
      }}
      \end{mdframed}
      \vspace{0.3cm}
    \item \textbf{User Age Prediction}: \\
      \begin{mdframed}[nobreak=true]
      \small {\texttt{Based on this information, what is my age range? Choose one from the following age ranges:\\
      Under 18, 18-24, 25-34, 35-44, 45-49, 50-55, 56+.\\
      Structure the answer with one age range only without extra words.\\
      Answer: 
      }}
      \end{mdframed}
      \vspace{0.3cm}
    \item \textbf{User Occupation Prediction}: \\
      \begin{mdframed}[nobreak=true]
      \small {\texttt{Based on this information, what is my occupation? Choose three from the following occupations:\\
      academic, artist, clerical, college student, customer service, health care, executive, farmer, homemaker, K-12 student, lawyer, programmer, retired, sales, scientist, self-employed, technician, craftsman, unemployed, writer, other.\\
      From the list above, select three occupations that are an exact match and structure the answer in the format of "OCCUPATION1$\rightarrow$OCCUPATION2$\rightarrow$\\OCCUPATION3".\\
      Answer:
      }}
      \end{mdframed}
      \vspace{0.3cm}
\end{itemize}

\subsubsection{Amazon Review\\}

All unseen prediction tasks for Amazon Review dataset share the same prompt preamble: 
\begin{tcolorbox}{
\texttt{Here is a summary of my interest about \{domain\} products based on my product review history (between \textless{}summary\textgreater{} and \textless{}/summary\textgreater{}).\\
\textless{}summary\textgreater{}SUMMARY\textless{}/summary\textgreater{}
}}\end{tcolorbox}

For each prediction task, the prompt suffix is as following:
\begin{itemize}
    \item \textbf{Favorite Category Prediction}: \\
      \begin{mdframed}[nobreak=true]
      \small {\texttt{Based on this information, what are my top 3 favorite categories of \{domain\}? Choose three from the following categories that are separated by "--\textgreater{}":\\
      \{choices\}\\
      From the list above, select three categories that are an exact match and structure the answer in the format of "CATEGORY1--\textgreater{}CATEGORY2--\textgreater{}CATEGORY3".\\
      Answer: 
      }}
      \end{mdframed}
      \vspace{0.3cm}
    \item \textbf{Rating Prediction}: \\ 
      \begin{mdframed}[nobreak=true]
      \small {\texttt{Based on this information, what is my rating on the following \{domain\} product? \\
      Answer a score between 1 to 5 inclusive, using full-point(1.0) increments.\\
      Structure the answer with "[Your choice from (1.0, 2.0, 3.0, 4.0, 5.0)]".\\
      Answer:
      }}
      \end{mdframed}
      \vspace{0.3cm}
    \item \textbf{Review Generation}:  \\
      \begin{mdframed}[nobreak=true]
      \small {\texttt{In addition, here is a list of few example reviews I wrote for each product in \{domain\}, in format of [PRODUCT\_NAME: "REVIEW\_TEXT"]:\\
      Based on this information, write a review based on my preference and writing style for the following product in \{domain\}: \\
      Structure the answer with "[Your review text for this product]".\\
      Answer:
      }}
      \end{mdframed}
      \vspace{0.3cm}
\end{itemize}

\subsubsection{Google Local Review\\}

All unseen prediction tasks for Google Local Review dataset share the same prompt preamble: 
\begin{tcolorbox}
\texttt{
Here is a summary of my place visit pattern and preference in \{state\} state based on my place visit history (between \textless{}summary\textgreater{} and \textless{}/summary\textgreater{}).\\
\textless{}summary\textgreater{}SUMMARY\textless{}/summary\textgreater{}
}
\end{tcolorbox}

For each prediction task, the prompt suffix is as following:
\begin{itemize}
    \item \textbf{Favorite Category Prediction}: \\
      \begin{mdframed}[nobreak=true]
      \small {\texttt{Based on this information, what are my top 3 favorite categories for places to visit in \{state\} state? Choose three from the following categories that are separated by "--\textgreater{}":\\
      \{choices\}\\
      From the list above, select three categories that are an exact match and structure the answer in the format of "CATEGORY1--\textgreater{}CATEGORY2--\textgreater{}CATEGORY3".\\
      Answer:
      }}
      \end{mdframed}
      \vspace{0.3cm}
    \item \textbf{Rating Prediction}:  \\
      \begin{mdframed}[nobreak=true]
      \small {\texttt{Based on this information, what is my rating on the following place in \{state\} state?\\
      Answer a score between 1 to 5 inclusive, using full-point(1.0) increments.\\
      Structure the answer with "[Your choice from (1.0, 2.0, 3.0, 4.0, 5.0)]".\\
      Answer:
      }}
      \end{mdframed}
      \vspace{0.3cm}
    \item \textbf{Common Activity(City) Retrieval}: \\
      \begin{mdframed}[nobreak=true]
      \small {\texttt{Based on this information, what are the top 3 cities I visited most in \{state\} state? Choose three from the following city names that are separated by "--\textgreater{}":\\
      \{choices\} \\
      From the list above, select three city names that are an exact match and structure the answer in the format of "CITY1--\textgreater{}CITY2--\textgreater{}CITY3".\\
      Answer:
      }}
      \end{mdframed}
      \vspace{0.3cm}
\end{itemize}

\subsection{Prompts for Few-Shot Summarization}
For each datasets, we randomly selected 2 examples outside of the evaluation set, and used Gemini 1.5 Pro to generate user summaries. The prompt used for example user summary generation is the same as RLPF Prompt in Table \ref{tab: summarization_prompts}.

\subsubsection{MovieLens2015\\}
\small{Question1: Here is a list of movies that I have watched in the format of [MOVIE\_NAME (YEAR), MY\_RATING], where RATING=1 is the lowest and RATING=5 is the highest: [Gothika (2003), 3.5], [Cat in the Hat, The (2003), 2.5], [Identity (2003), 4.0], [Scary Movie (2000), 2.0], [Scary Movie 2 (2001), 2.0], [Queen of the Damned (2002), 2.0], [Jeepers Creepers (2001), 2.0], [Scary Movie 3 (2003), 2.0], [Ghost Ship (2002), 2.0], [Book of Shadows: Blair Witch 2 (2000), 2.0], [FearDotCom (a.k.a. Fear.com) (a.k.a. Fear Dot Com) (2002), 2.0], [Elf (2003), 4.0], [Matrix Revolutions, The (2003), 3.0], [Holes (2003), 4.0], [Fast and the Furious, The (2001), 3.0], [Mission: Impossible II (2000), 3.0], [xXx (2002), 4.0], [Matrix Reloaded, The (2003), 3.0], [Catch Me If You Can (2002), 3.0], [Lara Croft Tomb Raider: The Cradle of Life (2003), 3.0], [Charlie's Angels (2000), 3.5], [Charlie's Angels: Full Throttle (2003), 3.5], [Adventures of Pluto Nash, The (2002), 2.5], [Tuxedo, The (2002), 3.0], [Showtime (2002), 3.0], [Legally Blonde (2001), 4.0], [Coming to America (1988), 4.0], [American President, The (1995), 4.0], [Christmas Story, A (1983), 4.5], [Mister Roberts (1955), 3.0], [Ferris Bueller's Day Off (1986), 4.5], [Harvey (1950), 3.0], [Monty Python Live at the Hollywood Bowl (1982), 4.5], [Stuck on You (2003), 3.0], [Cheaper by the Dozen (2003), 3.0], [League of Extraordinary Gentlemen, The (a.k.a. LXG) (2003), 3.0], [Chicago (2002), 3.0], [Chitty Chitty Bang Bang (1968), 3.5], [Maverick (1994), 3.0], [Titanic (1997), 4.5], [Mask of Zorro, The (1998), 4.0], [Top Gun (1986), 4.0], [Patriot, The (2000), 3.0], [Atlantis: The Lost Empire (2001), 3.0], [Wild Thornberrys Movie, The (2002), 3.0], [Laputa: Castle in the Sky (TenkÃ83Â no shiro Rapyuta) (1986), 4.0], [Simone (S1m0ne) (2002), 3.0], [Seabiscuit (2003), 3.5], [On Golden Pond (1981), 3.0], [Legally Blonde 2: Red, White \& Blonde (2003), 3.0]. Please summarize my movie watch preference within 200 words.

\noindent Answer: Based on your provided list and ratings, you seem to enjoy a wide variety of movie genres.  You've watched numerous comedies, action films, horror/thrillers, and family movies, with a few classic films sprinkled in.  While there's a concentration of releases from the early 2000s, indicating a potential fondness for movies from that era, your list also includes older films spanning several decades.    You appear to enjoy popular franchises like "Scary Movie," "Charlie's Angels," and "The Matrix," suggesting an appreciation for sequels and familiar concepts. However, your inclusion of critically acclaimed films like "Chicago" and "Seabiscuit," alongside lighter fare like "Cat in the Hat," indicates that your taste isn't limited to blockbusters.  Further analysis with more data and specific ratings could refine this summary and reveal more nuanced preferences.  For example, providing your ratings would help determine which genres you truly prefer versus those you simply watch. \\

\noindent Question2: Here is a list of movies that I have watched in the format of [MOVIE\_NAME (YEAR), MY\_RATING], where RATING=1 is the lowest and RATING=5 is the highest: [Fletch (1985), 4.0], [EDtv (1999), 4.0], [Liar Liar (1997), 3.0], [Peggy Sue Got Married (1986), 3.0], [Ace Ventura: When Nature Calls (1995), 3.0], [Dear God (1996), 3.0], [Drop Dead Fred (1991), 1.0], [Pallbearer, The (1996), 3.0], [Fatal Instinct (1993), 1.0], [High School High (1996), 1.0], [U2: Rattle and Hum (1988), 5.0], [Madonna: Truth or Dare (1991), 2.0], [Taxi Driver (1976), 4.0], [Searching for Bobby Fischer (1993), 5.0], [Mr. Holland's Opus (1995), 3.0], [Boogie Nights (1997), 3.0], [Dead Poets Society (1989), 5.0], [American History X (1998), 3.0], [Philadelphia (1993), 3.0], [Gattaca (1997), 4.0], [Rain Man (1988), 5.0], [Contact (1997), 4.0], [Simon Birch (1998), 5.0], [Twelve Monkeys (a.k.a. 12 Monkeys) (1995), 4.0], [Field of Dreams (1989), 4.0], [Primary Colors (1998), 2.0], [2001: A Space Odyssey (1968), 4.0], [Truman Show, The (1998), 4.0], [Rob Roy (1995), 3.0], [What's Eating Gilbert Grape (1993), 2.0], [Quiz Show (1994), 4.0], [Karate Kid, The (1984), 5.0], [Crimson Tide (1995), 3.0], [Kalifornia (1993), 3.0], [Color of Money, The (1986), 3.0], [Amistad (1997), 5.0], [Firm, The (1993), 3.0], [He Got Game (1998), 2.0], [Phenomenon (1996), 5.0], [Pink Floyd: The Wall (1982), 4.0], [Interview with the Vampire: The Vampire Chronicles (1994), 3.0], [Othello (1995), 4.0], [Rainmaker, The (1997), 3.0], [Last Days of Disco, The (1998), 3.0], [Cook the Thief His Wife \& Her Lover, The (1989), 1.0], [Rudy (1993), 4.0], [Swing Kids (1993), 3.0], [Robin Hood: Prince of Thieves (1991), 4.0], [Horse Whisperer, The (1998), 4.0], [What Dreams May Come (1998), 2.0]. Please summarize my movie watch preference within 200 words.

\noindent Answer: Based on your viewing history, you enjoy a diverse range of films, spanning comedies, dramas, biopics, and science fiction. You seem to appreciate 90s cinema, with a particular interest in films exploring social issues, personal journeys, and human relationships.  Comedies like "Liar Liar" and "Ace Ventura" suggest a taste for slapstick humor, while "Dead Poets Society," "Gattaca," and "Rain Man" indicate a preference for thought-provoking dramas with strong emotional cores.  You also enjoy historical dramas like "Amistad" and "Rob Roy," and classics like "Taxi Driver" and "2001: A Space Odyssey." While you have watched several critically acclaimed movies, your ratings also reveal an openness to less acclaimed films, showing an appreciation for diverse storytelling, regardless of critical reception. Your inclusion of music documentaries like "U2: Rattle and Hum" suggests an interest in music and artist biographies.  Overall, you appear to be a movie enthusiast with broad tastes, enjoying both popular and critically acclaimed films across various genres. \\

\noindent Question3: \{input\_activities\}}

\subsection{Amazon Review\\}

\small{Question1: Here is a list of Books products that I have purchased and ratings I gave for each product in the format of [PRODUCT\_NAME, MY\_RATING], where MY\_RATING=1 is the lowest and MY\_RATING=5 is the highest: [The Unearthly (The Unearthly Series) (Volume 1), 4.0], [Swallow the Moon, 4.0], [Harbored Secrets, 4.0], [Risen (The Wintergrave Chronicles) (Volume 2), 5.0], [Ebola K: A Terrorism Thriller (Volume 1), 5.0], [You Never Called Me Princess, 4.0], [Arkadia: The Halfway House Series (Volume 1), 5.0], [12.21.12, 5.0], [The Game, 4.0], [Tur: The Elsker Saga (Volume 1), 3.0], [Home Grown (Modern Contemporary Fiction), 4.0], [Love at First Date (Better Date than Never), 3.0], [The Contradiction of Solitude, 5.0], [Blood of Requiem: Song of Dragons, Book 1, 4.0], [The Execution (The Wintergrave Chronicles) (Volume 1), 5.0], [The Noru: Blue Rose (Volume 1), 4.0], [Roomies, 5.0], [The Experiment of Dreams, 4.0], [30 Days, 4.0], [Dying to Forget (The Station) (Volume 1), 4.0], [Now or Never, 4.0], [The Trouble With Goodbye (A Fairhope New Adult Romance) (Volume 1), 5.0], [ The Watchtower eBook, 1.0], [Raven (HalfwayHouse Series) (Volume 2), 5.0], [Within: A Medical Suspense Novel, 3.0], [When We Were Dragons, 2.0], [ The Woodsman eBook, 3.0], [The Vampire Reclamation Project (Volume 1), 4.0], [Safe and Sound, 5.0], [ Betrayal, 5.0], [Grade A Stupid: Book 1 of the Darcy Walker Series, 4.0], [BROWN BREAD: an exceptionally humorous literary satire - Kindle edition, 4.0], [Elisha's Bones, 4.0], [The Forgotten Ones (The Danaan Trilogy) (Volume 1), 4.0], [Feral: Book 1 in the Werecat Saga, 3.0], [A Death Displaced: Lansin Island Series (Volume 1), 4.0], [ Nathanial\&39;s Window, 4.0], [Dusk (Magic) (Volume 1), 4.0], [Leashed (Going to the Dogs) (Volume 1), 3.0], [Crave the Rose, 3.0], [Terminal Rage, 5.0], [The Lonely, 4.0], [The Good Life, 4.0], [Eyes of Osiris: A Kayara Ingham Trilogy Novel (The Architechts of Lore) (Volume 1), 5.0], [ The Deathtaker, 5.0], [Sending Jack Off To Jesus (A Southern Thing) (Volume 2), 5.0], [The Housewife Assassin's Handbook (The Housewife Assassin Series) (Volume 1), 4.0], [Blood and Lilies (Bloodlines), 3.0], [Without, 5.0], [Jury of Peers, 5.0]. Please summarize my interest about Books within 200 words.

\noindent Answer: Based on your purchases and ratings, you enjoy a wide range of genres within Books.  Thrillers and suspense are prominent, with titles like "Ebola K," "Within," and "Terminal Rage." Fantasy and paranormal also appeal to you, as evidenced by "The Unearthly," "Blood of Requiem," "The Vampire Reclamation Project," "Feral," and "Dusk (Magic)." Romance features in your list with books like "Love at First Date," "The Trouble With Goodbye," and "Crave the Rose." You also explore contemporary fiction ("Home Grown"), literary satire ("BROWN BREAD"), and mysteries ("A Death Displaced," "Eyes of Osiris"). Series books, particularly in fantasy and thriller genres, seem to catch your eye, suggesting an interest in immersive worlds and continuing storylines. Finally, the inclusion of "The Housewife Assassin's Handbook" hints at an appreciation for dark humor or unconventional narratives. \\

\noindent Question2: Here is a list of Books products that I have purchased and ratings I gave for each product in the format of [PRODUCT\_NAME, MY\_RATING], where MY\_RATING=1 is the lowest and MY\_RATING=5 is the highest: [Search for the Buried Bomber (Dark Prospects), 3.0], [Forty Times a Killer:: A Novel of John Wesley Hardin (Bad Men of the West), 3.0], [The Fringe Worlds: Book 1 of The Human Chronicles Saga (Volume 1), 4.0], [Breakfast at Sally's: One Homeless Man's Inspirational Journey, 5.0], [Sihpromatum: I Grew My Boobs in China (Volume 1), 1.0], [The Normans: From Raiders to Kings, 3.0], [Ancient Aliens In Australia: Pleiadian Origins of Humanity, 4.0], [Lost in Shangri-La: Escape from a Hidden World - A True Story, 3.0], [Lost in Shangri-La: Escape from a Hidden World, 3.0], [Michael Jackson Conspiracy, 1.0], [Napoleon's Hemorrhoids: And Other Small Events That Changed the World, 2.0], [Georgia O'Keeffe Museum, 5.0], [Confessions Of A Gunfighter, 4.0], [The Ghost Files (The Ghost Files (Book 1)), 1.0], [The Prepper's Grid Down Survival Guide: How To Prepare If The Lights Go Out \&amp; The Gas, Water Or Electricity Grid Collapses, 4.0], [The Shopkeeper (Steve Dancy Tale), 4.0], [Cherokee Clans: An Informal History (Cherokee Chapbooks 4), 5.0], [ A Letter from Auschwitz eBook, 5.0], [Jungleland: A Mysterious Lost City and a True Story of Deadly Adventure (P.S.), 5.0], [Skinwalker Ranch: Path of the Skinwalker, 4.0], [Lost on Skinwalker Ranch: The True Story of a Property Guard and His Encounter with the Paranormal, 5.0], [Tales Behind the Tombstones: The Deaths And Burials Of The Old Wests Most Nefarious Outlaws, Notorious Women, And Celebrated Lawmen, 2.0], [Plans For Time Travel Machines That Really Work - Revised And Updated Edition: How To Move Through Time And Space, 4.0], [The Lost Tribe of Coney Island: Headhunters, Luna Park, and the Man Who Pulled Off the Spectacle of the Century, 3.0], [Dead Angler (Loon Lake Fishing Mystery), 2.0], [Scenic Driving New Mexico, 3.0], [The Devil's Grin - A Crime Novel featuring Anna Kronberg and Sherlock Holmes, 3.0], [Native American Survival Skills, 4.0], [Leadville (Center Point Western), 4.0], [Voices from Legendary Times: We are a bridge between past and future, 4.0], [ Rupee Millionaires eBook, 5.0], [Kevin and I in India (Book 2 of 6 in the Frank's Travel Memoir Serie) (Volume 2), 5.0], [Parsons Green, 3.0], [Ghosts of Gettysburg, III: Spirits, Apparitions and Haunted Places of the Battlefield, Vol. 3, 4.0], [The Oracle Code (Thomas Lourds), 5.0], [The Oath of Nimrod: Giants, MK-Ultra and the Smithsonian Coverup (Templars America), 2.0], [On Killing: The Psychological Cost of Learning to Kill in War and Society, 5.0], [Where the West Ends: Stories from the Middle East, the Balkans, the Black Sea, and the Caucasus, 3.0], [The American heritage history of the great West,, 4.0], [Virtue, Valor, and Vanity: The Inside Story of the Founding Fathers and the Price of a More Perfect Union, 4.0], [The Time Ships, 5.0], [Jesse James and the Secret Legend of Captain Coytus, 1.0], [Georgia O'Keeffe: American Art Series, 5.0], [Georgia O'Keeffe, 1887-1986: Flowers in the desert, 5.0], [The World to Come: The Guides' Long-Awaited Predictions for the Dawning Age, 4.0], [Aliens Among Us, 5.0], [The Practice and Science of Drawing (Dover Art Instruction), 3.0], [Magicians of the Gods: Sequel to the International Bestseller Fingerprints of the Gods, 4.0], [Sacred Activations: 26 Essential Sacred Activations To Expand Your Gifts Beyond Mastery, 1.0], [Selected by Extraterrestrials: My life in the top secret world of UFOs, think-tanks and Nordic secretaries, 5.0]. Please summarize my interest about Books within 200 words.

\noindent Answer: Based on your reading history, you have a strong interest in non-fiction, particularly in history, true crime, and the paranormal.  You're fascinated by historical figures like Napoleon and Jesse James, and enjoy learning about historical events, especially those with a focus on the American West and Native American culture.  A significant portion of your choices explore mysterious phenomena, such as aliens, UFOs, time travel, and unexplained events like those at Skinwalker Ranch. You also enjoy reading survival guides and books about travel, especially those with a personal touch or an adventurous element.  Finally, you seem drawn to art, with a particular interest in Georgia O'Keeffe. Overall, your reading preferences suggest a curious mind that enjoys exploring a range of factual topics with a leaning towards the unusual and unexplained. \\

\noindent Question3: \{input\_activities\}}

\subsubsection{Google Local Review\\}

\small{Question1: Here is a list of places that I have visited in New York state and ratings I gave for each place in the format of [PLACE\_NAME, CITY\_NAME, MY\_RATING], where MY\_RATING=1 is the lowest and MY\_RATING=5 is the highest:  [Country Cottage, 5.0], [(716) Food and Sport, 4.0], [Target, 4.0], [Lockport Cave \& Underground Boat Ride, 3.0], [Lake Effect Ice Cream, 5.0], [Wegmans, 5.0], [Chipotle Mexican Grill, 4.0], [Lake Erie Grape Discovery Center, 4.0], [Pace's Pizzeria, 4.0], [Cross Roads Farm \& Craft Market, 4.0], [Mississippi Mudds, 4.0], [Niawanda Park, 4.0], [Lakewood Beach, 3.0], [Tim Hortons, 4.0], [Tops Friendly Markets, 3.0], [Dash's Market, 5.0], [Tim Hortons, 4.0], [The Village Casino, 3.0], [Panama Rocks Scenic Park, 4.0], [Transit Drive-In Theatre, 5.0], [Walmart Supercenter, 4.0], [PetSmart, 4.0], [Wegmans, 5.0], [Tim Hortons, 3.0], [Royal Family Restaurant, 4.0], [Dairy Queen (Treat), 4.0], [McDonald's, 3.0], [Santora's Pizza Pub \& Grill, 4.0], [Pepe's Pizzeria, 4.0], [Tim Hortons, 4.0], [Kohl's, 4.0], [Walmart Supercenter, 4.0], [Wiechec's Lounge, 4.0], [Anderson's Frozen Custard, 5.0], [Ted's Hot Dogs, 5.0], [Kenmore Fish Market, 4.0], [Best Buy, 4.0], [Tops Friendly Markets, 3.0], [Shannon Pub, 4.0], [The Dockside Bar \& Grill, 4.0], [Ellicott Creek Park, 4.0], [Old Man River, 4.0], [Old Man River, 4.0], [Santora's Pizza Pub \& Grill, 4.0], [VCA Sheridan Animal Hospital, 4.0], [Average Joe's, 4.0], [Polish Villa II, 4.0], [Polish Villa II, 4.0], [Mr. Tire Auto Service Centers, 4.0], [Tim Hortons, 4.0]. Please summarize my location history pattern and place visit preference within 200 words.

\noindent Answer: Based on your location history, you appear to spend a significant amount of time in Western New York, potentially near Buffalo or Niagara Falls, given mentions of places like Niawanda Park, Lakewood Beach, the Lockport Cave, and the Lake Erie Grape Discovery Center.  You frequently visit major chains like Wegmans, Tops, Tim Hortons, Walmart, McDonald's, and Target, suggesting a preference for convenience and familiarity. You also enjoy local restaurants, with pizza places (Santora's, Pepe's, Pace's) being particularly frequent, and show an interest in local attractions and recreation, visiting parks (Ellicott Creek, Panama Rocks), markets (Cross Roads, Dash's, Kenmore Fish Market), and entertainment venues (The Village Casino, Transit Drive-In).  No ratings were provided, so specific place preferences based on quality can't be determined, but the frequency of visits implies a general satisfaction with these establishments. The inclusion of VCA Sheridan Animal Hospital suggests you may be a pet owner.\\   

\noindent Question2: Here is a list of places that I have visited in New York state and ratings I gave for each place in the format of [PLACE\_NAME, CITY\_NAME, MY\_RATING], where MY\_RATING=1 is the lowest and MY\_RATING=5 is the highest:  [Super 8 by Wyndham Kenmore/Buffalo/Niagara Falls Area, 5.0], [Life Storage - Buffalo, 5.0], [Fashion Outlets of Niagara Falls USA, 5.0], [Metro by T-Mobile, 5.0], [Seneca Buffalo Creek Casino, 5.0], [River Grill, 5.0], [Erie Basin Marina, 5.0], [Tim Hortons, 5.0], [UPS Customer Center, 5.0], [Tim Hortons, 5.0], [Tim Hortons, 5.0], [Metro by T-Mobile, 5.0], [Chuck E. Cheese, 5.0], [Carbone's Pizza, 5.0], [Audiomotive Creations, 5.0], [McDonald's, 5.0], [Bon-Ton, 5.0], [Supercuts, 5.0], [Days Inn by Wyndham Tonawanda/Buffalo, 5.0], [Kenmore Village Restaurant, 5.0], [Red Lobster, 5.0], [McDonald's, 5.0], [Joe's New York Style Pizza, 5.0], [Joe's New York Style Pizza, 5.0], [MLK Jr. Park, 5.0], [MLK Jr. Park, 5.0], [Mississippi Mudds, 5.0], [Tops Friendly Markets, 5.0], [Tim Hortons, 5.0], [Tim Hortons, 5.0], [Smoke on the Water, 5.0], [Five Below, 5.0], [Del-Ton Plaza, 5.0], [Union Consumer Square, 5.0], [Hoag's Gas Station \& Smoke Shop, 5.0], [Tim Hortons, 5.0], [Queen City Wine \& Liquor, 5.0], [Catch 22 Bar and Grill, 5.0], [Wegmans, 5.0], [Save A Lot, 5.0], [gulf, 5.0], [Tops Elmwood Plaza, 5.0], [Tops Elmwood Plaza, 5.0], [Tim Hortons, 5.0], [La Nova, 5.0], [Hair World Beauty Supply, 5.0], [Tops Friendly Markets, 5.0], [Save A Lot, 5.0], [Delaware Consumer Square, 5.0], [Public Storage, 5.0]. Please summarize my location history pattern and place visit preference within 200 words.

\noindent Answer: Your New York state location history suggests a concentration in the Buffalo/Niagara Falls area, with visits spanning various establishment types. You frequent chain establishments like Tim Hortons, McDonald's, Tops Friendly Markets, and Save A Lot, indicating a preference for convenience and familiarity.  Your visits to Fashion Outlets of Niagara Falls USA and various plazas (Union Consumer Square, Delaware Consumer Square, Tops Elmwood Plaza) suggest an interest in shopping and retail.  You also visited several restaurants (River Grill, Carbone's Pizza, Kenmore Village Restaurant, Red Lobster, Joe's New York Style Pizza, Catch 22 Bar and Grill, La Nova), demonstrating a diverse culinary exploration.  Inclusion of entertainment venues like Seneca Buffalo Creek Casino and Chuck E. Cheese, alongside parks (MLK Jr. Park) indicates a balance of leisure activities.  However, the absence of specific ratings prevents a deeper understanding of your true preferences and satisfaction levels for these locations. \\

\noindent Question3: \{input\_activities\}}

\end{document}